\documentclass[pdflatex,sn-mathphys-num]{sn-jnl}% Math and Physical Sciences Numbered Reference Style
\usepackage{graphicx}%
\usepackage{multirow}%
\usepackage{amsmath,amssymb,amsfonts}%
\usepackage{amsthm}%
\usepackage{mathrsfs}%
\usepackage[title]{appendix}%
\usepackage{xcolor}%
\usepackage{textcomp}%
\usepackage{manyfoot}%
\usepackage{booktabs}%
\usepackage{algorithm}%
\usepackage{algorithmicx}%
\usepackage{algpseudocode}%
\usepackage{listings}%
\usepackage{array}
\usepackage{subfigure}
\usepackage{natbib}
\usepackage{hyperref} % for the journal
\usepackage{caption}
\theoremstyle{thmstyleone}%
\theoremstyle{thmstyletwo}%

\theoremstyle{thmstylethree}%

\begin{document}

\title[Article Title]{Hierarchical place recognition with omnidirectional images and curriculum learning-based loss functions}

%%=============================================================%%
%% GivenName	-> \fnm{Joergen W.}
%% Particle	-> \spfx{van der} -> surname prefix
%% FamilyName	-> \sur{Ploeg}
%% Suffix	-> \sfx{IV}
%% \author*[1,2]{\fnm{Joergen W.} \spfx{van der} \sur{Ploeg} 
%%  \sfx{IV}}\email{iauthor@gmail.com}
%%=============================================================%%

\author*[1]{\fnm{Marcos} \sur{Alfaro}}\email{malfaro@umh.es}

\author[1]{\fnm{Juan José} \sur{Cabrera}}\email{juan.cabreram@umh.es}

\author[1]{\fnm{María} \sur{Flores}}\email{m.flores@umh.es}

\author[1,2]{\fnm{Oscar} \sur{Reinoso}}\email{o.reinoso@umh.es}

\author[1,2]{\fnm{Luis} \sur{Payá}}\email{lpaya@umh.es}

\affil*[1]{\orgdiv{Institute for Engineering Research (I3E)}, \orgname{Miguel Hernández University}, \orgaddress{\street{Av. Universidad, s/n}, \city{Elche}, \postcode{03202}, \country{Spain}}}

\affil[2]{\orgname{Camí de Vera, Building 3Q}, \orgaddress{\street{Street}, \city{Valencia}, \postcode{46020}, \country{Spain}}}

%%==================================%%
%% Sample for unstructured abstract %%
%%==================================%%

\abstract{This paper addresses Visual Place Recognition (VPR), which is essential for the safe navigation of mobile robots. The solution we propose employs panoramic images and deep learning models, which are fine-tuned with triplet loss functions that integrate curriculum learning strategies. By progressively presenting more challenging examples during training, these loss functions enable the model to learn more discriminative and robust feature representations, overcoming the limitations of conventional contrastive loss functions. After training, VPR is tackled in two steps: coarse (room retrieval) and fine (position estimation). The results demonstrate that the curriculum-based triplet losses consistently outperform standard contrastive loss functions, particularly under challenging perceptual conditions. To thoroughly assess the robustness and generalization capabilities of the proposed method, it is evaluated in a variety of indoor and outdoor environments. The approach is tested against common challenges in real operation conditions, including severe illumination changes, the presence of dynamic visual effects such as noise and occlusions, and scenarios with limited training data. The results show that the proposed framework performs competitively in all these situations, achieving high recognition accuracy and demonstrating its potential as a reliable solution for real-world robotic applications. The code used in the experiments is available at \url{https://github.com/MarcosAlfaro/TripletNetworksIndoorLocalization.git}.}

\keywords{place recognition, omnidirectional images, deep neural networks, triplet loss, curriculum learning}

%%\pacs[JEL Classification]{D8, H51}

%%\pacs[MSC Classification]{35A01, 65L10, 65L12, 65L20, 65L70}

\maketitle

\section{Introduction}\label{sec1}

The advancement of autonomous systems, from mobile robots to self-driving vehicles, relies on their capacity for robust and accurate localization within complex and dynamic environments. Place recognition, which involves capturing sensory information from the robot unknown position and retrieving the most similar sample among a previously stored database, is crucial for enabling this autonomy \cite{yin2025, ullah2024, bachute2021}.

\vspace{0.5cm}

Among various sensing modalities, vision sensors offer a compelling balance of rich information, capturing colors, textures or shapes, at a relatively low cost \cite{masone2021}. However, the utility of visual data is often challenged by perceptual aliasing (i.e. different places appearing similar) or significant changes in appearance caused by variations of illumination, weather, season or viewpoint \cite{yadav2023}. Omnidirectional cameras, which have a 360$^\circ$ field of view, constitute a particularly potent solution for Visual Place Recognition (VPR) \cite{amoros2020}. These sensors capture complete information from the environment regardless of the robot orientation, thereby mitigating viewpoint-dependent appearance shifts and providing comprehensive scene context. Such panoramic views can be achieved through various means, including multi-camera systems \cite{kneip2013}, catadioptric setups \cite{lin2021} or paired fisheye cameras \cite{flores2022}.

\vspace{0.5cm}

In recent years, deep learning tools have had a major impact on the field of mobile robotics \cite{mumuni2022}. In this context, Convolutional Neural Networks (CNNs) \cite{lecun1998} and subsequently Vision Transformers (ViTs) \cite{dosovitskiy2021} have demonstrated unparalleled success in learning powerful and discriminative image representations directly from data \cite{arandjelovic2016}.

\vspace{0.5cm}

To effectively train these deep networks, specialized architectures and learning strategies are crucial. Siamese \cite{yin2019} and triplet \cite{liu2017} networks, which employ multiple weight-sharing branches, enable the generation of a vast number of training samples from a limited set of images. This is crucial for learning fine-grained similarities and differences between places. This learning is guided by a loss function, which must be carefully chosen to compare network outputs with ground truth relationships \cite{musgrave2020}. Within this context, contrastive learning has proven particularly effective \cite{ren2023}. It consists in training the model to map similar (positive) inputs to nearby points in an embedding space and dissimilar (negative) inputs to distant points. For VPR, positive pairs can be images from the same location, while negative pairs are from distinct locations. Triplet networks, utilizing anchor, positive, and negative samples, are especially suitable for refining descriptor spaces for high-precision retrieval. 

\vspace{0.5cm}

Despite these advancements, effectively leveraging omnidirectional imagery for fine-grained, robust VPR across diverse conditions and with efficiency remains a significant challenge. Furthermore, optimizing the learning process itself is necessary for high-accuracy position retrieval.

\vspace{0.5cm}

Consequently, this paper addresses these challenges by proposing a novel hierarchical VPR framework that leverages panoramic images with deep networks, specifically fine-tuned through a two-stage process: an initial stage for coarse retrieval followed by a second stage for fine localization within the retrieved area. A key innovation of our approach lies in the introduction of novel contrastive triplet loss functions that explicitly incorporate curriculum learning principles. By strategically presenting training examples in an order of increasing difficulty, these loss functions guide the network to progressively learn more discriminative features, enhancing the final performance in the place recognition task. The experimental evaluations, conducted across challenging indoor and outdoor datasets, demonstrate that this approach, even with low-shot fine-tuning, achieves superior precision and remarkable robustness against substantial environmental changes. The main contributions of this paper are threefold:

\begin{itemize}
    \item We develop a framework to tackle hierarchical VPR with omnidirectional images and deep networks for coarse-to-fine place recognition.
    \item We propose a set of novel contrastive triplet loss functions that integrate curriculum learning strategies, dynamically adjusting the difficulty of training triplets to optimize feature embedding for VPR.
    \item We provide extensive experimental validation demonstrating that the proposed methodology provides a great precision and robustness against pronounced environmental changes, defying visual phenomena and different indoor and outdoor scenarios.
\end{itemize}

\vspace{0.5cm}

The rest of the manuscript is structured as follows. Section \ref{sec:sec2} reviews the state-of-the-art in VPR and relevant deep learning techniques. Section \ref{sec:sec3} presents the proposed hierarchical approach and novel loss functions. Section \ref{sec:sec4} describes the experimental setup and presents a comprehensive evaluation of the proposed framework. Finally, conclusions and future works are outlined in Section \ref{sec:sec5}.

\section{State of the art}\label{sec:sec2}

Visual Place Recognition (VPR) has long been an integral part of robot navigation and scene understanding, with early efforts focusing on analytical techniques to craft global appearance descriptors from images \cite{paya2018}. However, these handcrafted methods often struggled with significant appearance variations and scalability. The rise of deep learning, driven by increased computational capabilities, has fundamentally reshaped the VPR landscape. Pioneer convolutional neural networks (CNNs), initially designed for tasks like object classification (e.g., LeNet \cite{lecun1998}, and later developed by deeper models such as AlexNet \cite{krizhevsky2017}, VGG \cite{simonyan2014} and GoogLeNet \cite{szegedy2015} on datasets like ImageNet \cite{deng2009}), provided powerful tools for learning visual features directly from data. More recently, Vision Transformers (ViTs) \cite{dosovitskiy2021} and their variants, including Swin-L \cite{liu2022} or DINOv2 \cite{oquab2023}, have emerged as highly effective feature extractors, often demonstrating superior performance and generalization in various vision tasks, including VPR. 

\vspace{0.5cm}

Vision sensors are widely employed for VPR due to their versatility and low cost \cite{zhang2021, schubert2023, kazerouni2022}. Among these, omnidirectional cameras are particularly advantageous for this task. Their 360$^\circ$ field of view ensures that, for a given robot position, the captured image contains comprehensive scene information irrespective of the robot's orientation, inherently mitigating viewpoint-dependent appearance shifts. Consequently, numerous approaches have leveraged omnidirectional views as a primary or sole source of information for robot localization and mapping tasks, including panoramic \cite{rostkowska2023, cabrera2024a} and fisheye views \cite{flores2022}. While offering richer contextual information, effectively processing and learning from the often distorted or high-dimensional nature of panoramic imagery remains an active area of research.

\vspace{0.5cm}

Current VPR research focuses on optimizing deep learning architectures and their training processes to learn discriminative and robust place representations. Transformer-based architectures, such as AnyLoc \cite{keetha2023}, have been proposed to achieve high generalization capabilities across diverse environments. Concurrently, methods like CosPlace \cite{berton2022} have focused on developing efficient and scalable training strategies for large-scale VPR, utilizing modified CNN backbones (e.g., VGG16 \cite{simonyan2014}). EigenPlaces \cite{berton2023} further refined such CNN-based approach to enhance the robustness against viewpoint changes. Significant effort has also been invested in designing sophisticated pooling mechanisms to aggregate features into comprehensive global descriptors, as seen in MixVPR \cite{alibey2023} and SALAD \cite{izquierdo2024}. 

\vspace{0.5cm}
Despite their successes, many of these state-of-the-art methods demonstrate optimal performance primarily with standard, narrow field-of-view images and often require extensive fine-tuning for deployment in new environments or for achieving highly accurate localization. Furthermore, their predominant focus on single-step global localization may not be optimal for all scenarios.

\vspace{0.5cm}

In contrast to single-step global localization, hierarchical (or coarse-to-fine) strategies are often more suitable, particularly in structured environments like indoor spaces \cite{cebollada2022, cabrera2024b}. These methods typically decompose the localization problem into multiple stages: first, identifying a broader area or room containing the query image, and subsequently, estimating a precise position within that retrieved area. This staged approach can improve efficiency by reducing the search space and can enhance accuracy by allowing specialized models for each stage. 

\vspace{0.5cm}

Besides, learning discriminative embeddings is crucial for VPR. Siamese and triplet architectures \cite{chen2022,aftab2022,leyvavallina2021} are widely employed for this, trained using loss functions that explicitly model similarity and dissimilarity. Triplet loss, first introduced in fields like face recognition \cite{wu2017, schroff2015,hermans2017}, is particularly predominant, aiming to obtain similar embeddings of images from similar places (anchor-positive pairs) while obtaining dissimilar ones of images from different places (anchor-negative pairs). 

\vspace{0.5cm}

While effective, the design of loss functions specifically tailored and optimized for VPR remains an active research direction. To further enhance the learning process, curriculum learning strategies have shown promise. Curriculum learning involves presenting training examples to the model in a structured manner, typically from easy to progressively more difficult instances. This can be achieved through strategic example selection \cite{buyuktas2021} or by designing the loss function itself to adapt to the learning stage, as demonstrated by the siamese contrastive curriculum loss in \cite{zhang2024}. Given that triplet networks often outperform siamese networks in complex metric learning tasks \cite{olid2018}, extending curriculum learning principles to triplet loss functions for robust VPR is a compelling task. 

\vspace{0.5cm}

In this paper, a novel hierarchical VPR framework is proposed, which uniquely leverages the comprehensive information from panoramic images within a two-stage deep learning pipeline. A core contribution is the introduction of a set of contrastive triplet loss functions that integrate curriculum learning principles, designed to enhance feature discriminability and training stability. By combining the strengths of omnidirectional vision, hierarchical localization and curriculum-guided deep metric learning, the proposed approach aims to deliver superior accuracy and robustness, particularly in challenging real-world environments requiring low-shot fine-tuning. 

\section{Methodology}\label{sec:sec3}

\subsection{Hierarchical localization} \label{sec:sec31}

This approach leverages a place recognition framework based on omnidirectional imagery and deep neural networks, which are trained by means of a coarse-to-fine method. The input to our system consists of panoramic and equirectangular images captured by mobile robots across different indoor and outdoor environments. The datasets, along with the corresponding ground truth coordinates $(x,y)$ for each image, are partitioned into distinct training and test sets. The training images and the descriptors computed for each of them form the visual model, against which query images are compared. This supervised setup allows for precise training and evaluation of the proposed VPR pipeline.

\vspace{0.5cm}

A two-stage hierarchical localization strategy is implemented, adapting and extending concepts from \cite{cebollada2022}. Each stage utilizes a triplet network architecture, where three identical network branches that share their weights process an anchor image ($I_a$), a positive image ($I_p$) and a negative image ($I_n$), respectively, computing a descriptor for each image. During inference (validation and testing), a query image is processed by the appropriately trained model to generate a global appearance descriptor $\vec{d}_{test} \in \mathbb{R}^{D}$ (where $D=512$ in the experiments). This descriptor is normalized and compared with the descriptors in the visual map using Euclidean distance or cosine similarity. The nearest neighbor in the descriptor space determines the recognized place or estimated position. The hierarchical process unfolds as follows:

\subsubsection{Stage 1: Coarse Localization (Room Retrieval)} \label{sec:sec321}

The objective of this stage is to identify the correct room or general area where the query image was captured. For this purpose, the network is trained using triplets where the anchor $I_a$ and positive $I_p$ images belong to the same room, while the negative image $I_n$ is from a different room. Triplet samples are chosen randomly following these criteria. This way, the network learns to produce room-discriminative descriptors. At inference, for a query image, its descriptor $\vec{d}_{test_{1}}$ is compared against a set composed of one representative descriptor for each room, \(\boldsymbol{{D}_{I_{r}}}=\left [ \vec{d}_{I_{r_{1}}}, \vec{d}_{I_{r_{2}}}, ...,\vec{d}_{I_{r_{M}}} \right ]\), where $M$ is the total number of rooms. Each $\vec{d}_{I_{r_{j}}}$ is the descriptor of a pre-selected representative image for room $j$ (i.e. the image captured closest to the room's geometric center).

\vspace{0.5cm}

The Euclidean distance between the query descriptor and the nearest representative descriptor $dist_{min1}$ provides the room initially considered the best prediction $Room_{pred1}$. Let $c_1$ be the confidence for this top prediction. If $c_1$ is below a threshold $h_1$ (set to 0.5), the second-best prediction is also considered. If the confidence $c_2$ for this second prediction is above a threshold $h_2$ (set to 0.1), both rooms are passed to the fine localization stage. Otherwise, only the top-predicted room is considered. Success at this stage is defined as the correct identification of the room containing the query image. Figure \ref{fig:hierarchicalLoc} (a) displays the general outline of the coarse step.

\subsubsection{Stage 2: Fine Localization (Intra-Room Positioning)} \label{sec:sec322}

This stage aims to estimate the precise robot coordinates within the room(s) identified in the coarse step. In this stage, a unique model is trained for the entire environment. For this intra-room training, positive pairs $(I_a, I_p)$  are defined as images captured within a specified metric distance $r_+$ (set to 0.4m), while negative pairs $(I_a, I_n)$ are images captured further apart than the threshold $r_-$ (set to 0.4m). To perform inference, for a query image and a previously retrieved room $k$ (or rooms $k_1$ and $k_2$) from the coarse step, its descriptor $\vec{d}_{test_{2}}$ is compared against all descriptors in the visual map for that room(s), \(\boldsymbol{{D}_{FineLoc}^{VM}}=\left [ \vec{d}_{1}^{\ VM}, \vec{d}_{2}^{\ VM}, ...,\vec{d}_{n}^{\ VM} \right ]\) where $n$ is the number of images in the visual map, built from the retrieved rooms.

\vspace{0.5cm}

\begin{figure}[!htb]%
    \centering
      \subfigure[]{\includegraphics[width=\linewidth]{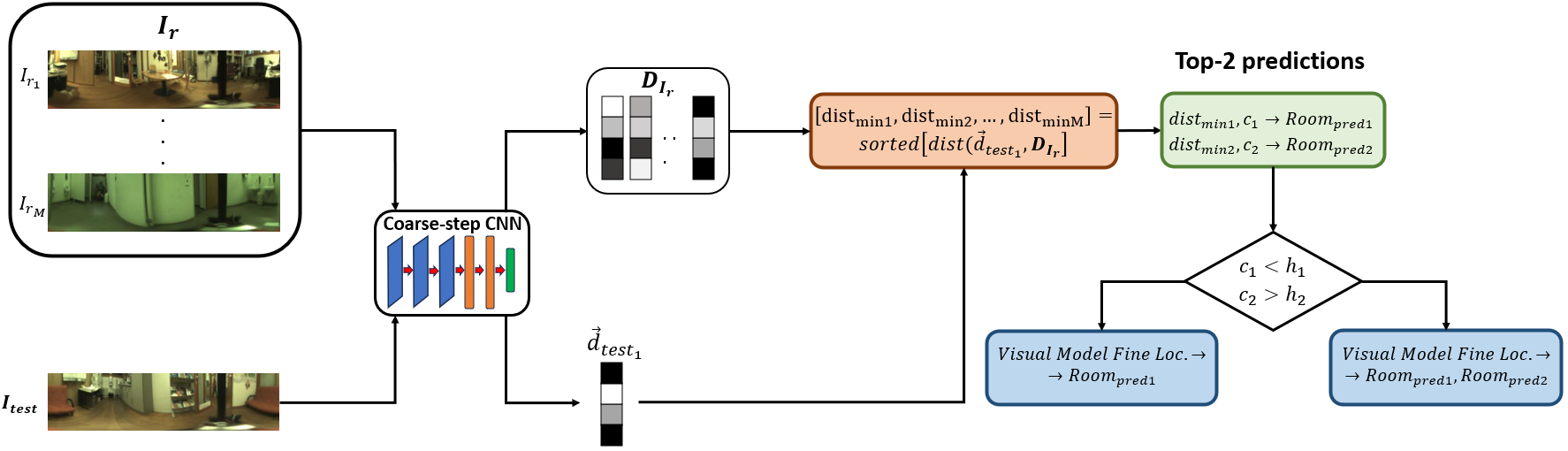}}
      \subfigure[]{ \includegraphics[width=\linewidth]{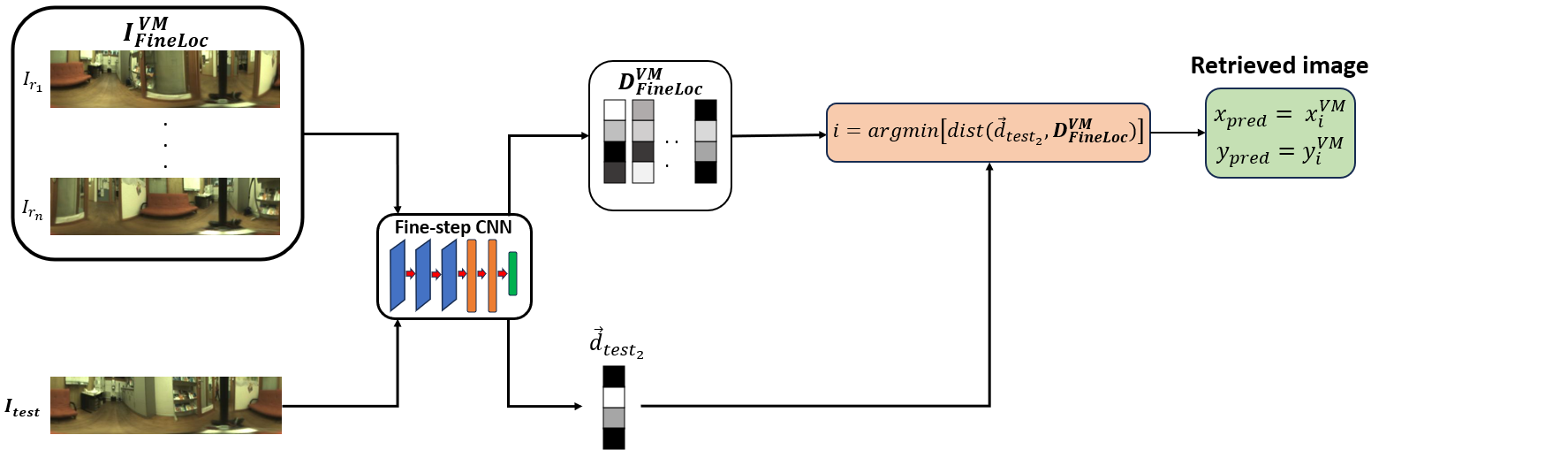}}
 \caption{Hierarchical localization process performed in two steps: (a) coarse localization (room retrieval); (b) fine localization (estimating the robot coordinates inside the retrieved room(s)).}
    \label{fig:hierarchicalLoc}
\end{figure}

The ground truth coordinates of the nearest neighbor in the visual map $(x_{i}^{VM}, y_{i}^{VM})$ are taken as the estimated robot position. Recall@1 ($R@1$) is typically measured as the percentage of queries correctly localized within a certain distance threshold $d$ from their true positions. Figure \ref{fig:hierarchicalLoc} (b) contains the scheme for the fine localization stage. 

\vspace{0.5cm}

\subsection{Backbone selection} \label{sec:sec32}

For both stages of the proposed hierarchical framework, we leverage the EigenPlaces \cite{berton2023} architecture, specifically the variant employing a VGG16 \cite{simonyan2014} backbone, which outputs a $D=512$ dimensional global descriptor. This choice was motivated by the fact that EigenPlaces demonstrated strong feature extraction capabilities and efficiency, with particular robustness to viewpoint changes, which is beneficial even with omnidirectional imagery. Preliminary experiments indicated its superiority over ResNet variants for this task. Transfer learning is employed by initializing the models with pre-trained weights and fine-tuning all the layers during the training process.

\subsection{Triplet loss functions} \label{sec:sec33}

\begin{table}[!htb]
\centering
\resizebox{\textwidth}{!}{
\small
\renewcommand{\arraystretch}{1.5}
\begin{tabular}{@{}>{\raggedright\arraybackslash}m{3.2cm} >{\centering\arraybackslash}m{8cm} >{\centering\arraybackslash}m{2cm}@{}}
\toprule
\textbf{Name} & \textbf{Formula} & \textbf{Parameters} \\
\midrule
Triplet Margin (TL) \cite{schroff2015} & 
$\mathcal{L} = \frac{1}{N}\sum_{i=1}^{N}[D_{a,p}^{i} - D_{a,n}^{i} + m]_{+}$ & 
$m$ \\
\addlinespace
Lifted Embedding (LE) \cite{hermans2017} & 
$\mathcal{L} = \frac{1}{N}\sum_{i=1}^{N} \left[ D_{a,p}^{i} + \ln\left(e^{m-D_{a,n}^{i}} + e^{m-D_{p,n}^{i}} \right) \right]_{+}$ & 
$m$ \\
\addlinespace
Lazy Triplet (LT) \cite{uy2018} & 
$\mathcal{L} = \left[ \max \left( \vec{D}_{a,p} - \vec{D}_{a,n} + m \right) \right]_{+}$ & 
$m$ \\
\addlinespace
Semi Hard (SH) \cite{schroff2015} & 
$\mathcal{L} = \frac{1}{N}\sum_{i=1}^{N} \left[ D_{a,p}^{i} - \min\left(\Vec{D}_{a,n}\right) + m \right]_{+}$ & 
$m$ \\
\addlinespace
Batch Hard (BH) \cite{hermans2017} & 
$\mathcal{L} = \left[ \max(\Vec{D}_{a,p}) - \min(\Vec{D}_{a,n}) + m \right]_{+}$ & 
$m$ \\
\addlinespace
Circle (CL) \cite{sun2020} & 
\shortstack{
$\mathcal{L} = \ln \left( 1 + \sum_{j=1}^{N} e^{\gamma \alpha _{n}^{j} s_{n}^{j}} + \sum_{i=1}^{N} e^{-\gamma \alpha _{p}^{i} s_{p}^{i}} \right)$ \\
$\text{where } \alpha_{p}^{i} = [O_{p} - s_{p}^{i}]_{+},\ \alpha_{n}^{j} = [s_{n}^{j} - O_{n}]_{+}$ \\
$O_{p} = 1 - m,\quad O_{n} = m$
} & 
$m$, $\gamma$ \\
\addlinespace
Angular (AL) \cite{wang2017} & 
\shortstack{
$\mathcal{L} = \ln\left( 1 + \sum_{i=1}^{N} e^{f_{a,p,n}^{i}} \right)$ \\
$\text{where } f_{a,p,n}^{i} = 4\tan^{2}\alpha(x_{a}^{i} + x_{p}^{i})^{T}x_{n}^{i}$ \\
$ - 2(1 + \tan^{2} \alpha)(x_{a}^{i})^{T}x_{p}^{i}$
} & 
$\alpha$ \\
\addlinespace
Curriculum TL+LT ($CV_{TL\xrightarrow{}LT}$) & 
$\mathcal{L} = w \cdot TL + (1-w) \cdot LT$ & 
$m_{TL}$, $m_{LT}$, $w$ \\
\addlinespace
Curriculum TL+BH ($CV_{TL\xrightarrow{}BH}$) & 
$\mathcal{L} = w \cdot TL + (1-w) \cdot BH$ & 
$m_{TL}$, $m_{BH}$, $w$ \\
\addlinespace
Curriculum LT+BH ($CV_{LT\xrightarrow{}BH}$) & 
$\mathcal{L} = w \cdot LT + (1-w) \cdot BH$ & 
$m_{LT}$, $m_{BH}$, $w$ \\
\bottomrule
\end{tabular}
}
\caption{Triplet loss functions employed in Experiment 1.}
\label{tab:table1}
\end{table}

\begin{table}[!htb]
\centering
\small
\resizebox{\textwidth}{!}{
\begin{tabular}{@{}c p{11cm}@{}}
\toprule
\textbf{Symbol} & \textbf{Description} \\
\midrule
$N$ & Batch size \\
\addlinespace
$x_{a}^{i}, x_{p}^{i}, x_{n}^{i}$ & Anchor, positive, and negative descriptors \\
\addlinespace
$D_{a,p}^{i}$ & Euclidean distance between the anchor and positive descriptors in the $i$-th triplet \\
\addlinespace
$D_{a,n}^{i}$ & Euclidean distance between the anchor and negative descriptors in the $i$-th triplet \\
\addlinespace
$D_{p,n}^{i}$ & Euclidean distance between the positive and negative descriptors in the $i$-th triplet \\
\addlinespace
$\vec{D}_{a,p}$ & Euclidean distances between each anchor-positive pair \\
\addlinespace
$\vec{D}_{a,n}$ & Euclidean distances between each anchor-negative pair \\
\addlinespace
$s_{p}^{i}$ & Cosine similarity between the anchor and positive descriptors in the $i$-th triplet \\
\addlinespace
$s_{n}^{j}$ & Cosine similarity between the anchor and negative descriptors in the $j$-th triplet \\
\addlinespace
$m$ & Margin \\
\addlinespace
$m_{TL}$, $m_{LT}$, $m_{BH}$ & Margin (TL,LT,BH) \\
\addlinespace
$\gamma$ & Scale factor (CL) \\
\addlinespace
$\alpha$ & Angular margin (AL) \\
\addlinespace
$w$ & Loss weight \\
\addlinespace
$[\cdot]_{+}$ & ReLU function: $\max(0, \cdot)$ \\
\bottomrule
\end{tabular}
}
\caption{Variables and parameters of the loss functions defined in Table \ref{tab:table1}.}
\label{tab:table2}
\end{table}

A core contribution of this paper is the exploration and proposal of novel triplet loss functions that integrate curriculum learning to enhance the training process for VPR. The standard triplet loss function aims to enforce a margin between the distance of positive pairs $(d(I_a, I_p))$ and negative pairs $(d(I_a, I_n))$. The loss functions that we propose in this work build upon this by introducing a curriculum strategy, which typically works by modulating the difficulty of triplets considered or by adjusting the loss emphasis as training progresses. This allows the network to first learn from easier examples and gradually tackle more challenging distinctions, leading to more robust and discriminative embeddings. To achieve this progressive training, two contrastive losses of different exigence are combined by means of a weighted sum:

\begin{equation}
   \mathcal{L}=\omega*\mathcal{L}_1+(1-\omega)*\mathcal{L}_2 
\end{equation}

\noindent where $\omega$ is the weight, $\mathcal{L}_1$ is the least restrictive loss and $\mathcal{L}_2$ is the hardest loss. 

This weight $\omega$ is initialized at 1 (i.e. the least restrictive loss is the only loss considered) and is modified during training, achieving a value of 0 at the end of the training process (i.e. only the hardest loss is employed). This approach permits performing a training with increasing difficulty, with the model first learning the broader features of the images from the dataset and finally adjusting to particular high-demanding examples to increase predictive accuracy. 

\vspace{0.5cm}

In this paper, three different curriculum loss variants are proposed, which are built by combining the Triplet Loss ($TL$) \cite{schroff2015}, which is calculated as the average error in the predictions from a batch, the Lazy Triplet ($LT$) \cite{uy2018}, which outputs the highest error from the batch (i.e. it returns the most difficult example from the batch), and the Batch Hard ($BH$) \cite{hermans2017}, which returns the most difficult positive and negative examples from the batch. In terms of training exigence, $TL<LT<BH$. Therefore, three different combinations are possible: $CV_{TL\xrightarrow{}LT}$, $CV_{TL\xrightarrow{}BH}$ and $CV_{LT\xrightarrow{}BH}$.

\vspace{0.5cm}

The specific formulations of the benchmark triplet losses and the proposed curriculum losses are detailed in Table \ref{tab:table1}. Definitions for all mathematical terms and symbols used in these loss functions are provided in Table \ref{tab:table2}. In the next section, exhaustive experiments will assess their impact on both coarse and fine localization performance.

\section{Experiments}\label{sec:sec4}

This section presents the experimental evaluation of the proposed hierarchical localization framework. Section \ref{sec:sec41} introduces the datasets used for training and evaluation, whereas the subsequent sections detail the experimental setup and results.

\subsection{Datasets} \label{sec:sec41}

\subsubsection{Indoor dataset: COLD}

The COLD database \cite{pronobis2009} is composed of omnidirectional images captured by a mobile robot that makes use of a catadioptric vision system with a hyperbolic mirror. The robot navigates various paths within several buildings, traversing different rooms such as offices, kitchens, toilets or corridors. The dataset was collected across four different scenarios --- Freiburg Part A (FR-A) and B (FR-B) and Saarbrücken Part A (SA-A) and B (SA-B) --- and under three different lighting conditions: cloudy, night and sunny.

\vspace{0.5cm}

\begin{figure}[!htb]%
    \centering
      \subfigure[]{\includegraphics[width=0.32\linewidth]{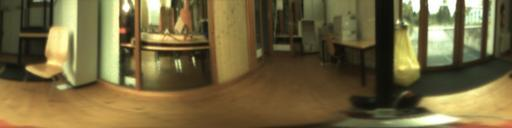}}
      \subfigure[]{ \includegraphics[width=0.32\linewidth]{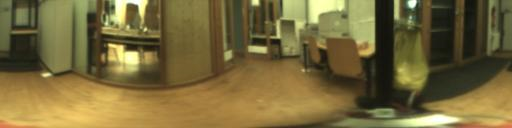}}
      \subfigure[]{\includegraphics[width=0.32\linewidth]{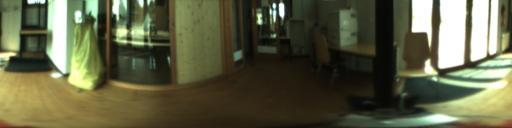}}
      \subfigure[]{\includegraphics[width=0.32\linewidth]{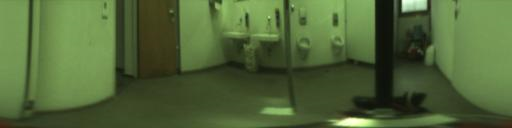}}
      \subfigure[]{ \includegraphics[width=0.32\linewidth]{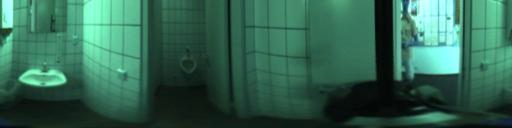}}
      \subfigure[]{\includegraphics[width=0.32\linewidth]{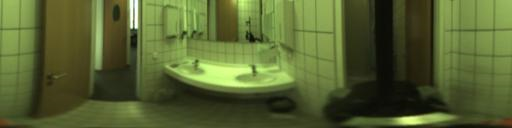}}
\caption{Sample images from the COLD database \cite{pronobis2009}. The top row shows different lighting conditions: (a) Cloudy, (b) Night, and (c) Sunny. The bottom row presents different environments: (d) FR-A, (e) SA-A, and (f) SA-B.}
\label{fig:imgExamples}
\end{figure}

Figure \ref{fig:imgExamples} displays sample images from the COLD database, illustrating the different lighting conditions and environments. These examples highlight challenging aspects of the dataset, such as significant appearance changes due to illumination variations (e.g. windows, shadows, etc) or pieces of furniture in different positions and perceptual aliasing arising from visually similar rooms in different locations.

\vspace{0.5cm}

\subsubsection{Mixed indoor-outdoor dataset: 360Loc}

\begin{figure}[!htb]%
    \centering
      \subfigure[]{\includegraphics[width=0.49\linewidth]{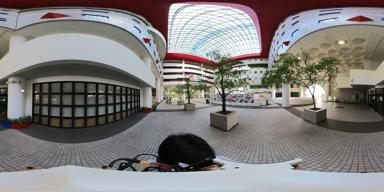}}
      \subfigure[]{ \includegraphics[width=0.49\linewidth]{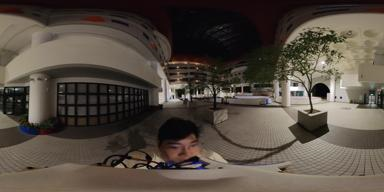}}
      \subfigure[]{\includegraphics[width=0.49\linewidth]{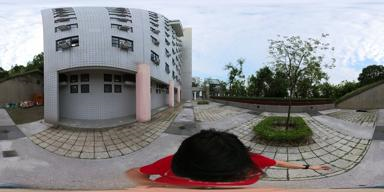}}
      \subfigure[]{\includegraphics[width=0.49\linewidth]{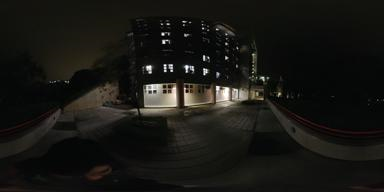}}
 \caption{Sample images from the 360Loc database \cite{huang2024}, captured in the (a, b) atrium and (c, d) hall environments under (a, c) day and (b, d) night conditions.}
    \label{fig:imgExamples360Loc}
\end{figure}

The 360Loc database \cite{huang2024} consists of equirectangular images captured with a 360$^\circ$ camera at the Hong Kong University, which includes both indoor and outdoor areas. This dataset contains trajectories from four different locations --- atrium, concourse, hall and piatrium --- recorded under both day and night conditions. Figure \ref{fig:imgExamples360Loc} shows examples of images from two different environments of this dataset (atrium and hall) under day and night illumination.

\subsection{Experiment 1. Evaluation of loss functions} \label{sec:sec42}

In this experiment, a comparative evaluation of the different triplet loss functions described in Section \ref{sec:sec3} is conducted in the FR-A environment. For both the coarse and fine localization stages, a separate network is trained with each loss function. We have varied the parameters of each loss function to identify their optimal configurations. Each network is trained with 50000 triplet samples, using the Stochastic Gradient Descent (SGD) as optimizer algorithm, with a learning rate $lr=0.001$.  Table \ref{tab:exp1imageSets} specifies the number of images used for training and testing in this experiment. Only cloudy images are used in the training. Tables \ref{tab:exp1CL} and \ref{tab:exp1FL} present the best results achieved with each loss function for coarse and fine localization, respectively. In this experiment, the threshold distance $d$ to calculate the $R@1$ was set to 0.5m.

\begin{table}[!htb]
  \centering
  % \resizebox{\linewidth}{!}{
  \begin{tabular}{lcccc}
    \toprule
    Environment & \begin{tabular}[c]{@{}c@{}}Train/ \\ Database (Cloudy) \end{tabular} & \begin{tabular}[c]{@{}c@{}}Test \\ Cloudy\end{tabular} & \begin{tabular}[c]{@{}c@{}}Test \\ Night\end{tabular} & \begin{tabular}[c]{@{}c@{}}Test \\ Sunny\end{tabular} \\
    \midrule
    FR-A & 556 & 2595 & 2707 & 2114 \\
    \bottomrule
  \end{tabular}
  % }
  \caption{Number of images from the FR-A environment used for training and evaluation in Experiment 1.}
  \label{tab:exp1imageSets}
\end{table}

\vspace{0.5cm}

\begin{table}[!htb]
\centering
\resizebox{\linewidth}{!}{
\begin{tabular}{lcccccc}
\toprule
\textbf{Loss} & \textbf{Parameters} & \textbf{N} & \multicolumn{4}{c}{\textbf{Accuracy (\%)}} \\
\cmidrule(lr){4-7}
 & & & \textbf{Cloudy} & \textbf{Night} & \textbf{Sunny} & \textbf{Avg.} \\
\midrule
TL            & $m=1$ & 16  & \textbf{99.27} & 97.41 & 95.70 & 97.46\\
LE            & $m=0.25$ & 4 & 99.15 & 97.38 & 95.65 & 97.39\\
LT            & $m=0.5$ & 4 & 99.19 & 97.27 & 96.03 & 97.50\\
SH            & $m=0.75$ & 16 & 99.00 & 97.19 & 95.70 & 97.30\\
BH            & $m=1$ & 4 & 98.96 & 97.27 & 96.12 & 97.45\\
CL            & $m=0.75,\gamma=1$ & 4 & 98.77 & 97.45 & 95.36 & 97.19\\
AL            & $\alpha=25^{\circ}$ & 16 & 98.96 & 97.08 & 95.79 & 97.28\\
$CV_{TL \rightarrow LT}$ & $m_{TL}=0.75, m_{LT}=1$ & 4 & 99.19 & 97.52 & 96.03 & 97.58\\
$CV_{TL \rightarrow BH}$ & $m_{TL}=0.75, m_{BH}=1$ & 4 & 99.19 & \textbf{97.60} & \textbf{96.36} & \textbf{97.72} \\
$CV_{LT \rightarrow BH}$ & $m_{LT}=0.5, m_{BH}=0.75$ & 8 & 99.11 & 97.56 & 96.26 & 97.64 \\
\bottomrule
\end{tabular}
}
\caption{Room retrieval accuracy (\%) for each loss function in the coarse localization task across different lighting conditions.}
\label{tab:exp1CL}
\end{table}

\vspace{0.5cm}

As shown in Table \ref{tab:exp1CL}, for coarse localization, the standard Triplet Loss (TL) achieved the highest accuracy when the test is performed in the same lighting conditions than the training (cloudy). However, the proposed curriculum-based loss functions demonstrated superior overall retrieval accuracy, particularly under lighting conditions not encountered during training (night and sunny). Specifically, the $CV_{TL\xrightarrow{}BH}$ loss variant yielded the best overall performance. 

\vspace{0.5cm}

\begin{table}[!htb]
\centering
\resizebox{\linewidth}{!}{
\begin{tabular}{lcccccc}
\toprule
\textbf{Loss} & \textbf{Parameters} & \textbf{N} & \multicolumn{4}{c}{\textbf{Recall@1 (\%)}} \\
\cmidrule(lr){4-7}
 & & & \textbf{Cloudy} & \textbf{Night} & \textbf{Sunny} & \textbf{Avg.} \\
\midrule
TL            & $m=0.75$ & 16  & 91.91 & 94.16 & 83.07 & 89.71\\
LE            & $m=0$ & 4 & 92.45 & 94.50 & 83.92 & 90.29\\
LT            & $m=0.5$ & 4 & 92.14 & \textbf{95.09} & 85.48 & 90.90\\
SH            & $m=0.75$ & 16 & \textbf{93.29} & 94.68 & 84.77 & 90.91\\
BH            & $m=0.75$ & 4 & 91.75 & 94.27 & 81.03 & 89.02\\
CL            & $m=1,\gamma=1$ & 4 & 92.91 & 94.50 & 84.67 & 90.69\\
AL            & $\alpha=30^{\circ}$ & 16 & 92.18 & 94.31 & 81.08 & 89.19\\
$CV_{TL \rightarrow LT}$ & $m_{TL}=0.5, m_{LT}=0.5$ & 4 & 93.14 & 94.87 & \textbf{85.71} & \textbf{91.24}\\
$CV_{TL \rightarrow BH}$ & $m_{TL}=0.5, m_{BH}=0.5$ & 4 & \textbf{93.29} & 94.57 & 83.35 & 90.40 \\
$CV_{LT \rightarrow BH}$ & $m_{LT}=0.5, m_{BH}=0.5$ & 8 & 92.41 & 94.64 & 84.96 & 90.67 \\
\multicolumn{3}{c}{Baseline (no training)} & 92.22 & 93.02 & 81.41 & 88.88 \\
\bottomrule
\end{tabular}
}
\caption{Recall@1 (\%) for each loss function in the fine localization task across different lighting conditions.}
\label{tab:exp1FL}
\end{table}

Table \ref{tab:exp1FL} indicates that for fine localization, the $CV_{TL\xrightarrow{}LT}$ loss function outperformed the other functions in terms of $R@1$, demonstrating its competitive performance across varied lighting scenarios. Besides, the Semi Hard (SH) and the Lazy Triplet (LT) losses also achieved high recall under cloudy and night conditions, respectively. When compared to the baseline, i.e. single-step localization without model fine-tuning is performed, the proposed training method based on curriculum loss functions showed a substantial improvement under all conditions, confirming its robustness against illumination changes.

\vspace{0.5cm}

\begin{figure}[!htb]%
    \centering
      \subfigure[]{\includegraphics[width=0.49\linewidth]{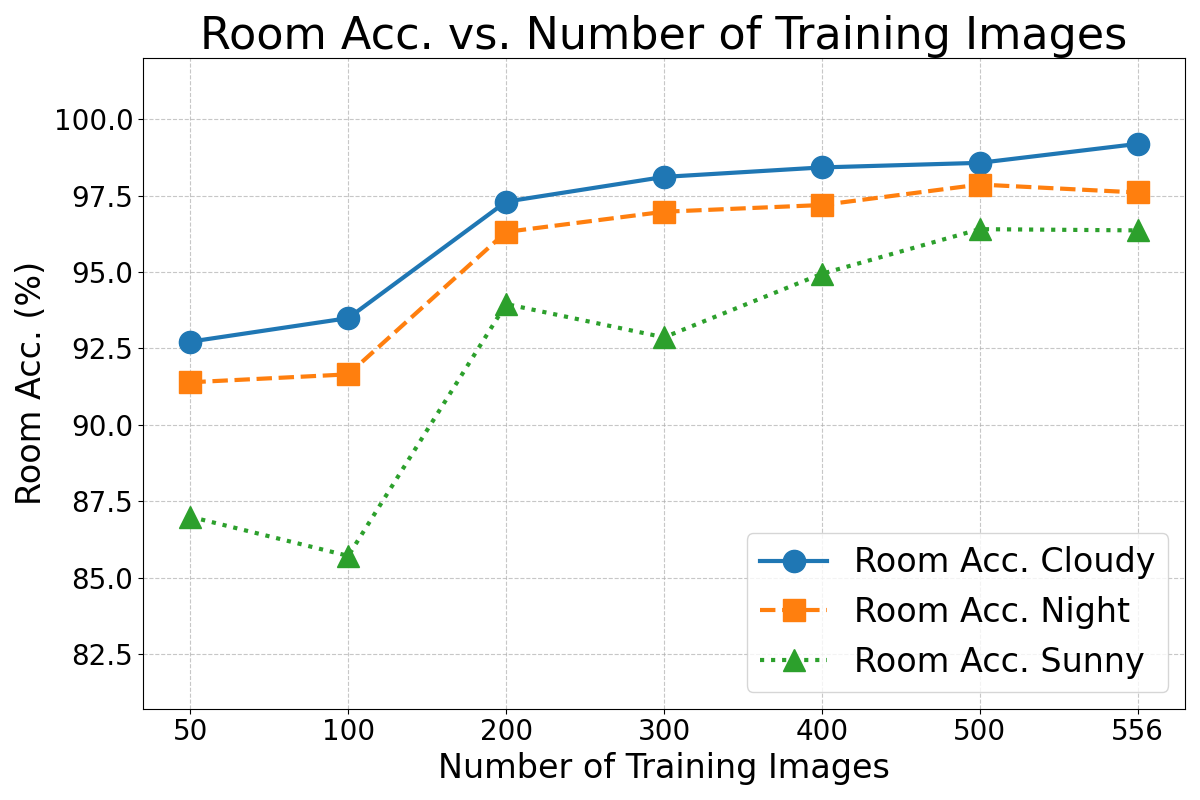}}
      \subfigure[]{\includegraphics[width=0.49\linewidth]{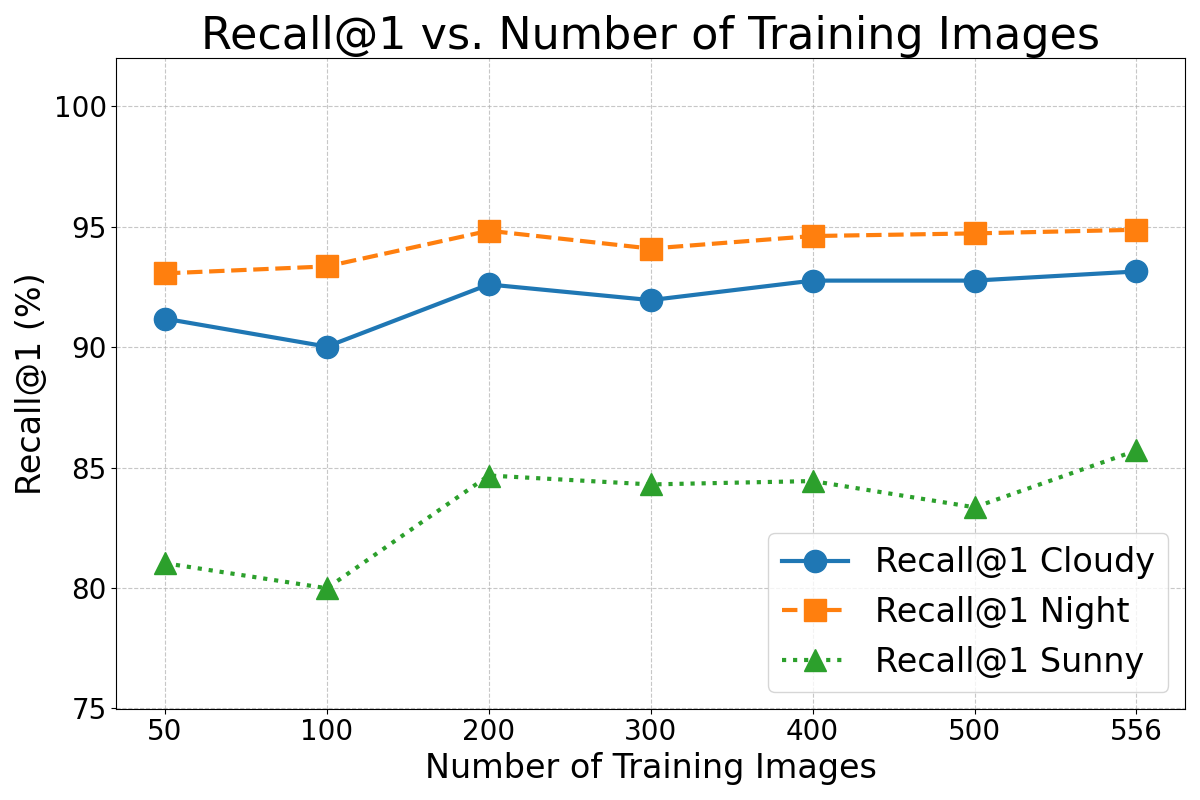}}
 \caption{Performance in (a) coarse localization and (b) fine localization for different training set sizes.}
    \label{fig:exp1numImages}
\end{figure}

Furthermore, the performance of the best loss function with varying sizes of the training set is evaluated. Figure \ref{fig:exp1numImages} (a) and (b) illustrate the results for each illumination condition as a function of the training set size.

The results in Figure \ref{fig:exp1numImages} indicate that the model performed competitively across all lighting conditions, even when trained on smaller subsets. The performance shows near-total invariance to the number of training images for sets with 200 or more images.

\vspace{0.5cm}

\begin{figure}[H]%
    \centering
      \subfigure[]{\includegraphics[width=0.48\linewidth]{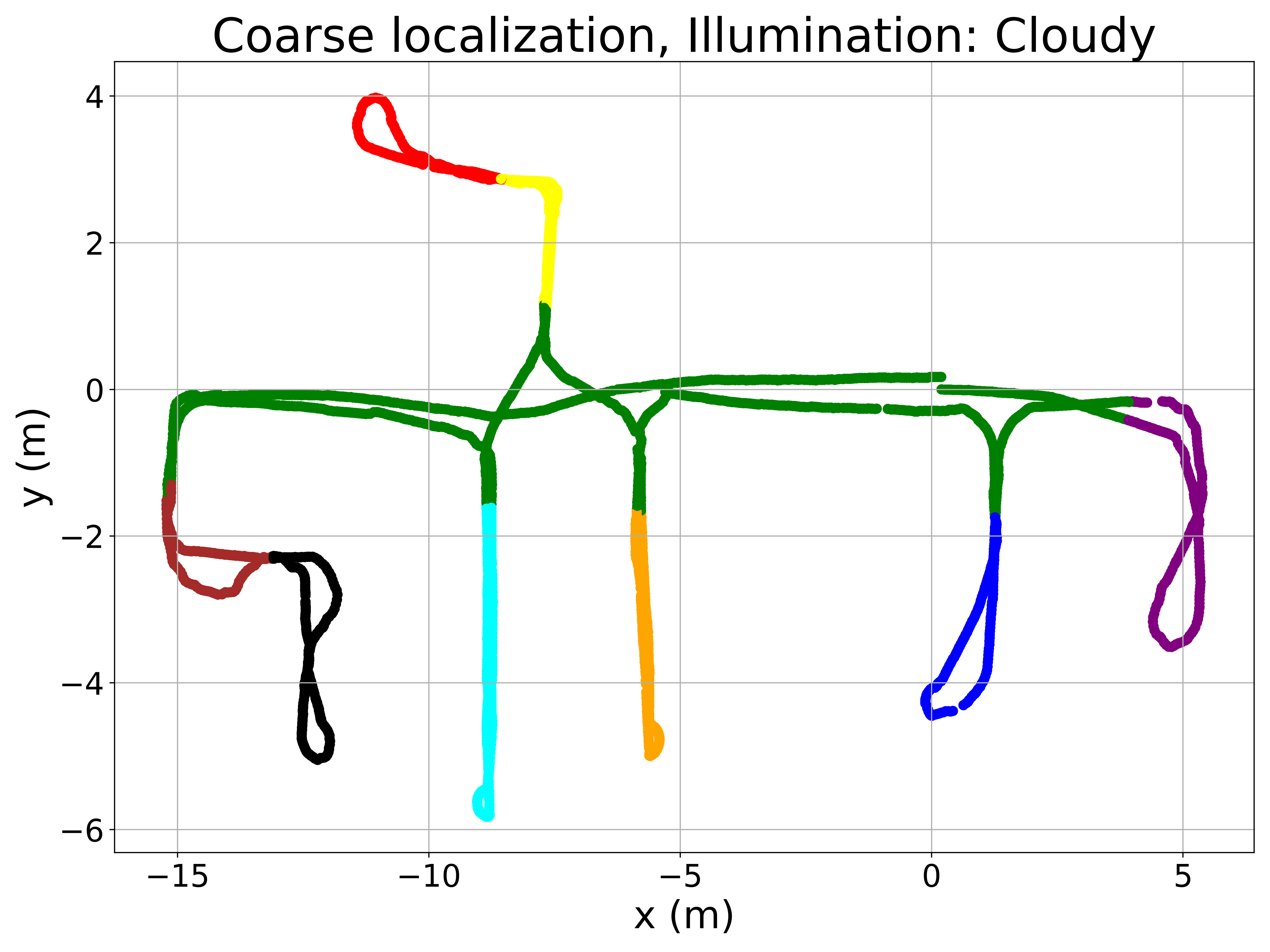}}
      \subfigure[]{\includegraphics[width=0.48\linewidth]{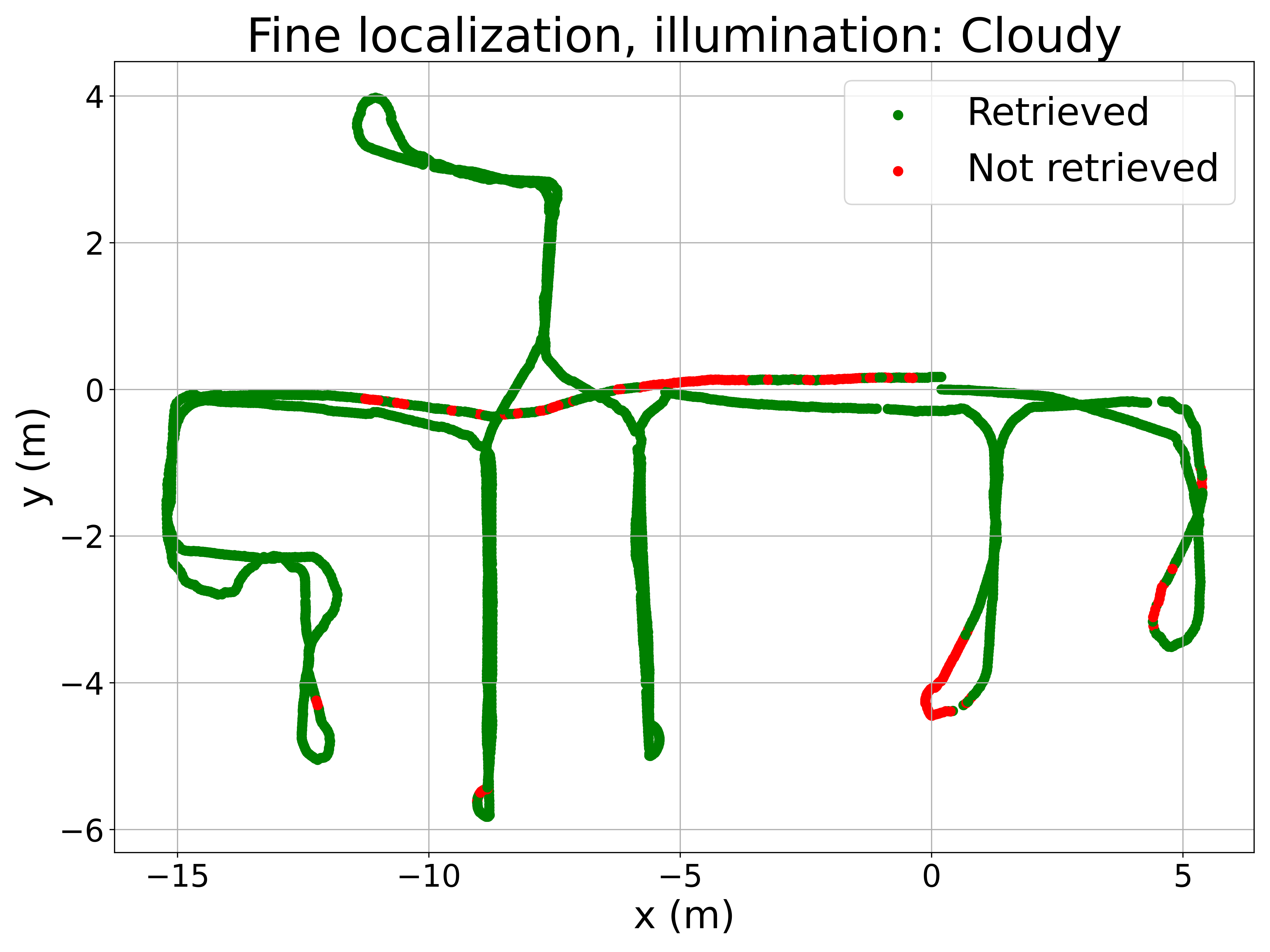}}
      \subfigure[]{\includegraphics[width=0.48\linewidth]{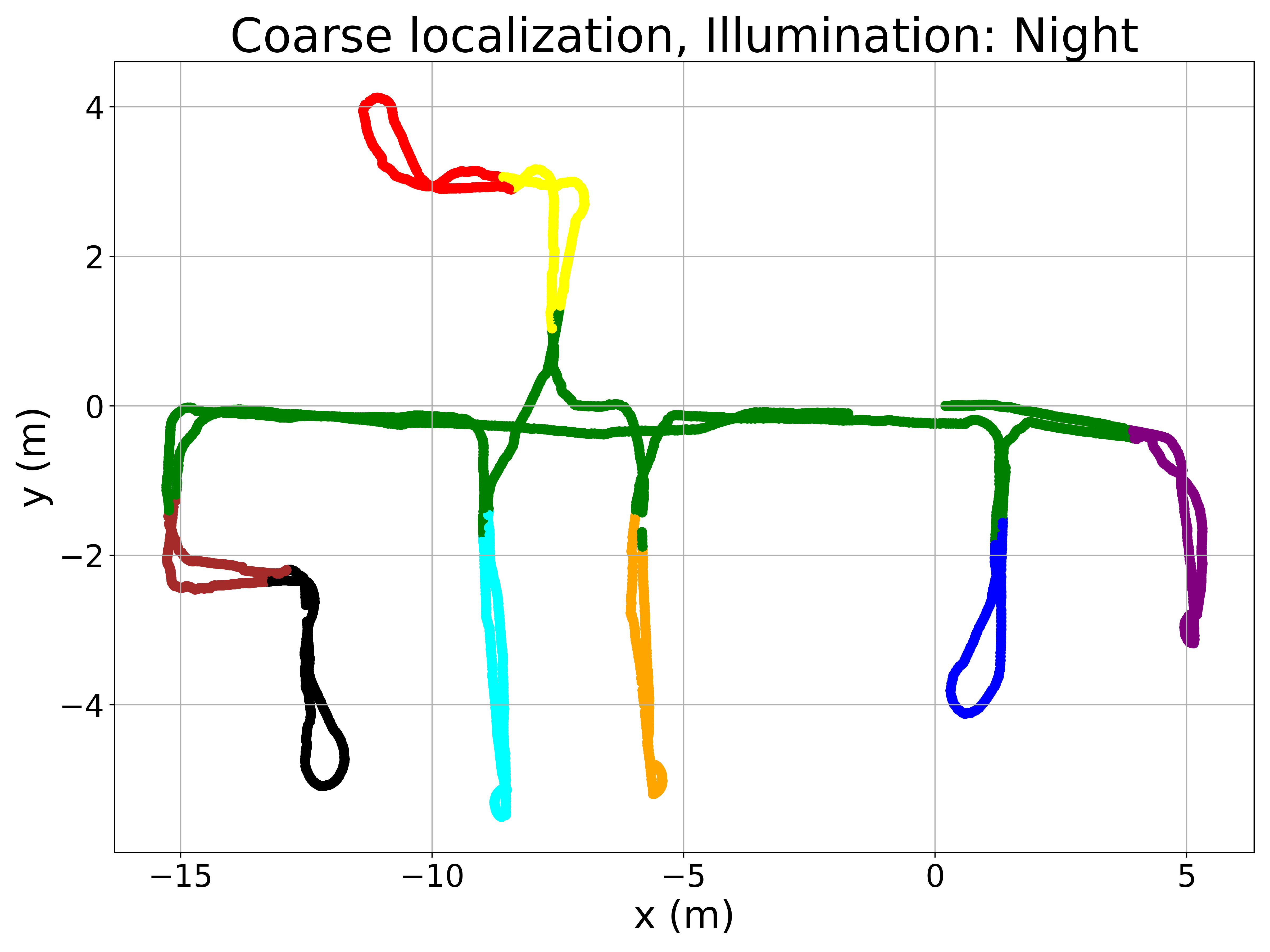}}
      \subfigure[]{\includegraphics[width=0.48\linewidth]{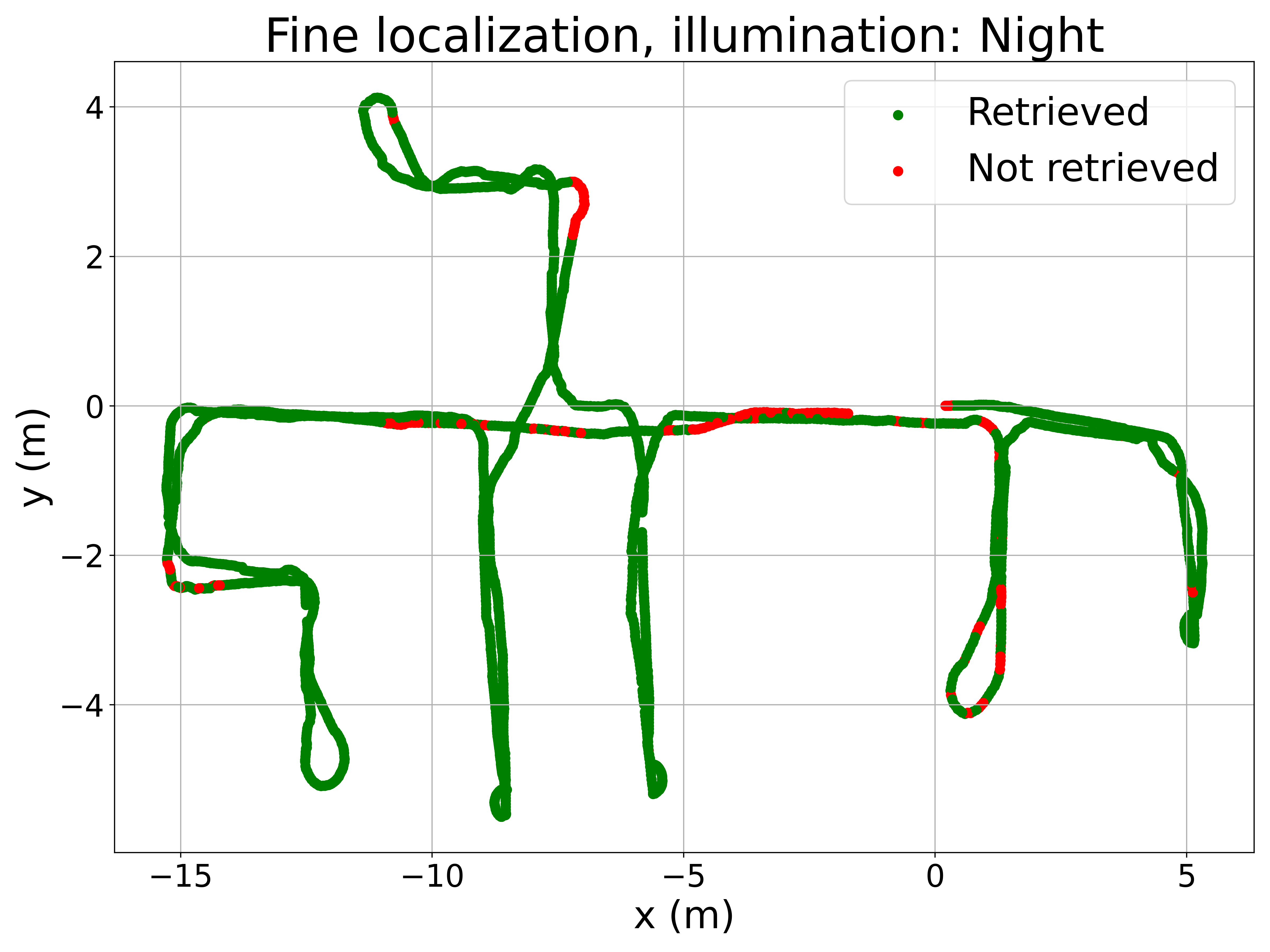}}
      \subfigure[]{\includegraphics[width=0.48\linewidth]{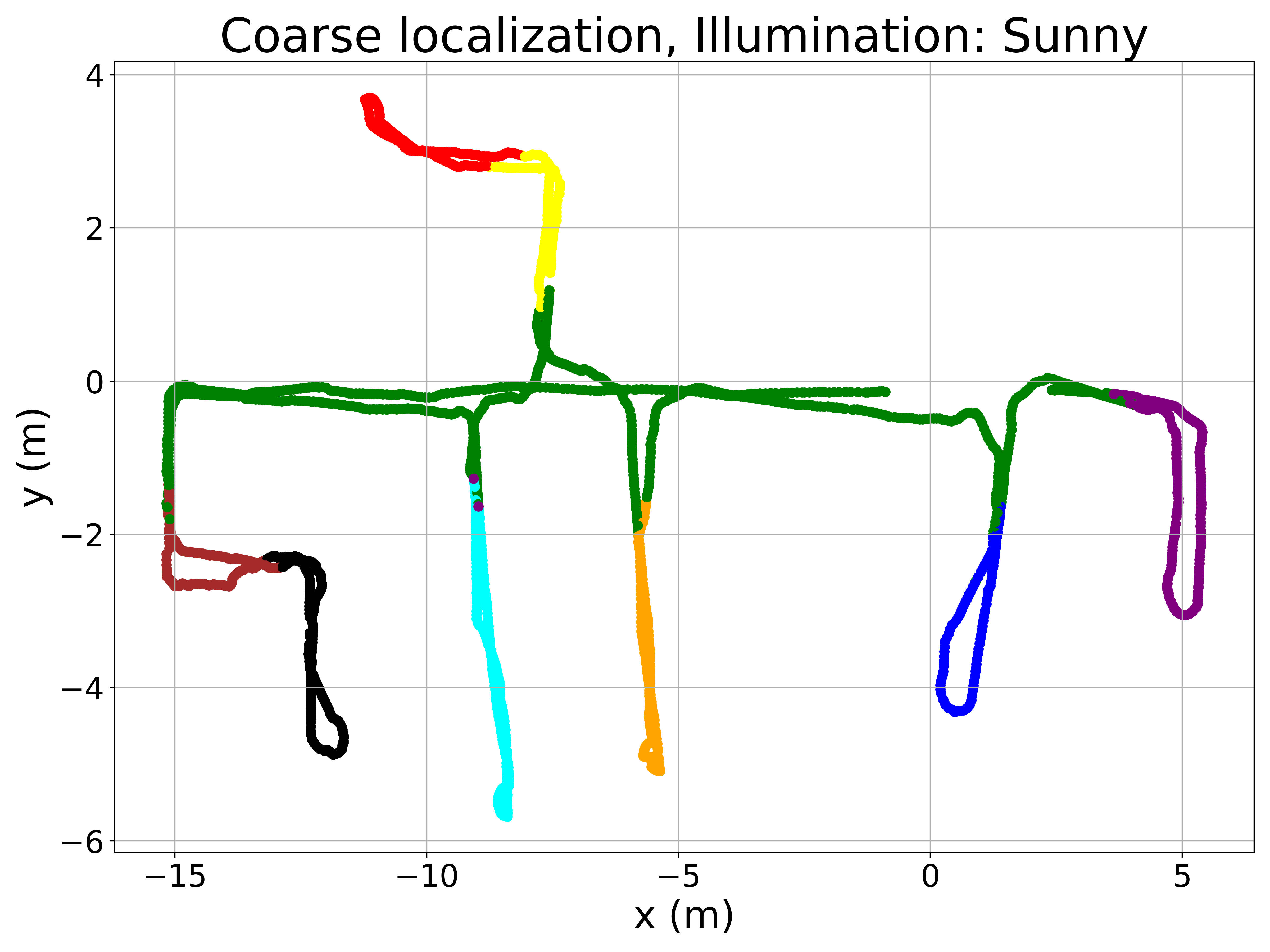}}
      \subfigure[]{\includegraphics[width=0.48\linewidth]{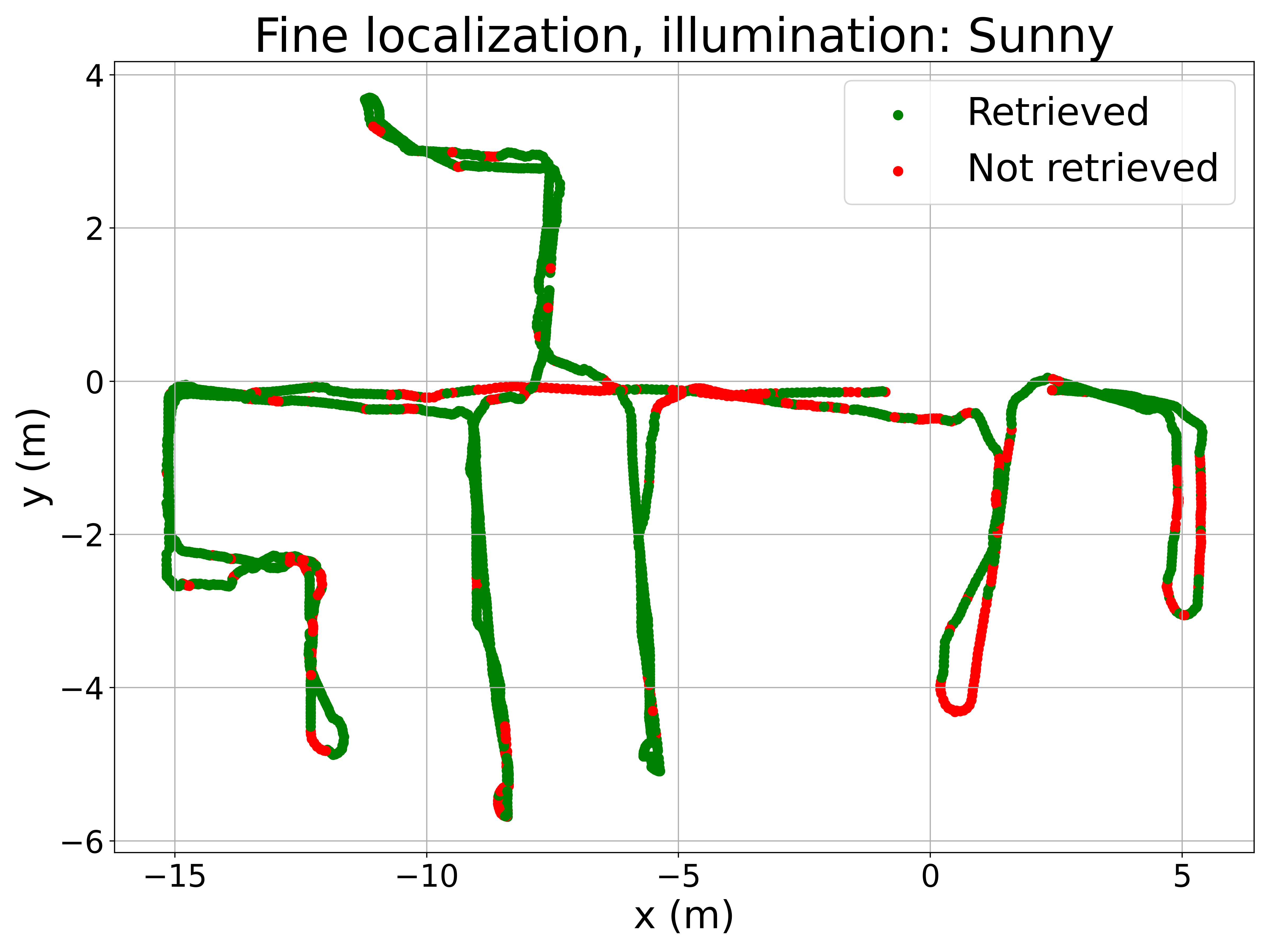}}
 \caption{Qualitative VPR results. The left column shows coarse localization (room retrieval) for (a) Cloudy, (c) Night, and (e) Sunny conditions. The right column shows fine localization (position retrieval) for (b) Cloudy, (d) Night, and (f) Sunny conditions.}
    \label{fig:exp1maps}
\end{figure}

To better understand the method's performance, Figure \ref{fig:exp1maps} presents qualitative results. The left column displays room retrieval predictions from the coarse model, where the query image locations are color-coded based on the retrieved room. The right column shows pose retrieval predictions from the fine model, where correctly localized query images are marked in green and incorrect ones in red. The Recall@1 criterium is used to consider correctly localized query images (i.e. the retrieved image must be at a distance lower than 0.5m from the query image). 

\vspace{0.5cm}

\vspace{0.5cm}

The maps in Figure \ref{fig:exp1maps} reveal a very low proportion of errors in room retrieval, considering that this is a challenging problem because of the visual similarity between different rooms. Errors between non-adjacent rooms are anecdotic, occurring only under sunny conditions. For fine localization, pose retrieval errors are concentrated in specific areas, such as the corridor, which is prone to perceptual aliasing. Some rooms with large windows also exhibit a higher number of errors under sunny conditions due to extreme illumination changes. Nevertheless, the proposed method demonstrates competitive performance across all lighting conditions, given the inherent difficulty of the task.

\vspace{0.5cm}

Regarding the computational cost, the localization time of the proposed hierarchical method, defined as the interval from image capture to the final pose retrieval, is approximately 7 ms. The memory requirement, which includes the descriptors for the query image and the visual map, as well as the two CNN models, is 1.72 GB. All experiments were conducted on a desktop computer equipped with an NVIDIA GeForce GTX 3090 GPU with 24 GB of RAM.

\subsection{Experiment 2. Analysis of the robustness against dynamic effects} \label{cap53}

\begin{figure}[!htb]%
    \centering
      \subfigure[]{\includegraphics[width=0.32\linewidth]{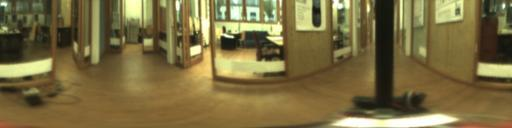}}
      \subfigure[]{\includegraphics[width=0.32\linewidth]{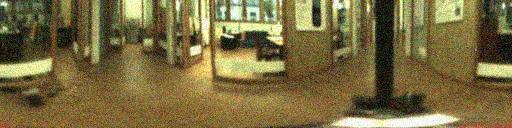}}
      \subfigure[]{\includegraphics[width=0.32\linewidth]{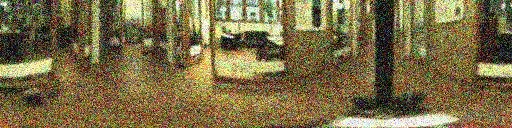}}
      \subfigure[]{\includegraphics[width=0.32\linewidth]{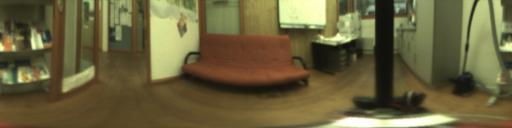}}
      \subfigure[]{\includegraphics[width=0.32\linewidth]{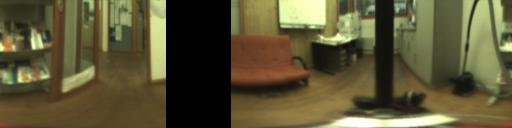}}
      \subfigure[]{\includegraphics[width=0.32\linewidth]{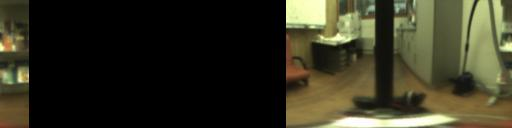}}
      \subfigure[]{\includegraphics[width=0.32\linewidth]{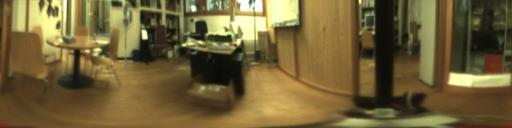}}
      \subfigure[]{\includegraphics[width=0.32\linewidth]{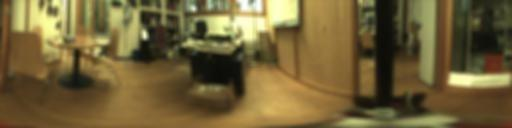}}
      \subfigure[]{\includegraphics[width=0.32\linewidth]{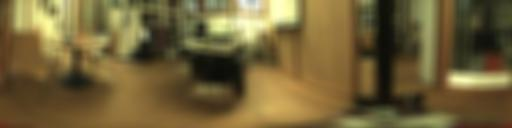}}
 \caption{Examples of panoramic images with different simulated effects. Left column: original images. Top row: Gaussian noise with (b) \(\sigma=20\) and (c) \(\sigma=50\). Middle row: occlusions of (e) 64 and (f) 128 columns. Bottom row: motion blur with a kernel size of (h) 7 and (i) 11 pixels.}
\label{fig:exp2effects}
\end{figure}

In this experiment, the robustness of the proposed hierarchical localization approach against common visual artifacts encountered in mobile robotics is assessed: occlusions, noise, and motion blur. Figure \ref{fig:exp2effects} provides examples of these effects applied to panoramic images from the FR-A environment. For this evaluation, the models trained for Experiment 1 are employed without any further training. The best-performing model from the coarse localization step $\left(CV_{TL\xrightarrow{}BH}\right)$ and the best model from the fine localization step $ \left( CV_{TL\xrightarrow{}LT} \right )$ are used.

\subsubsection{Noise effect}

\begin{figure}[!htb]%
    \centering
      \subfigure[]{\includegraphics[width=0.49\linewidth]{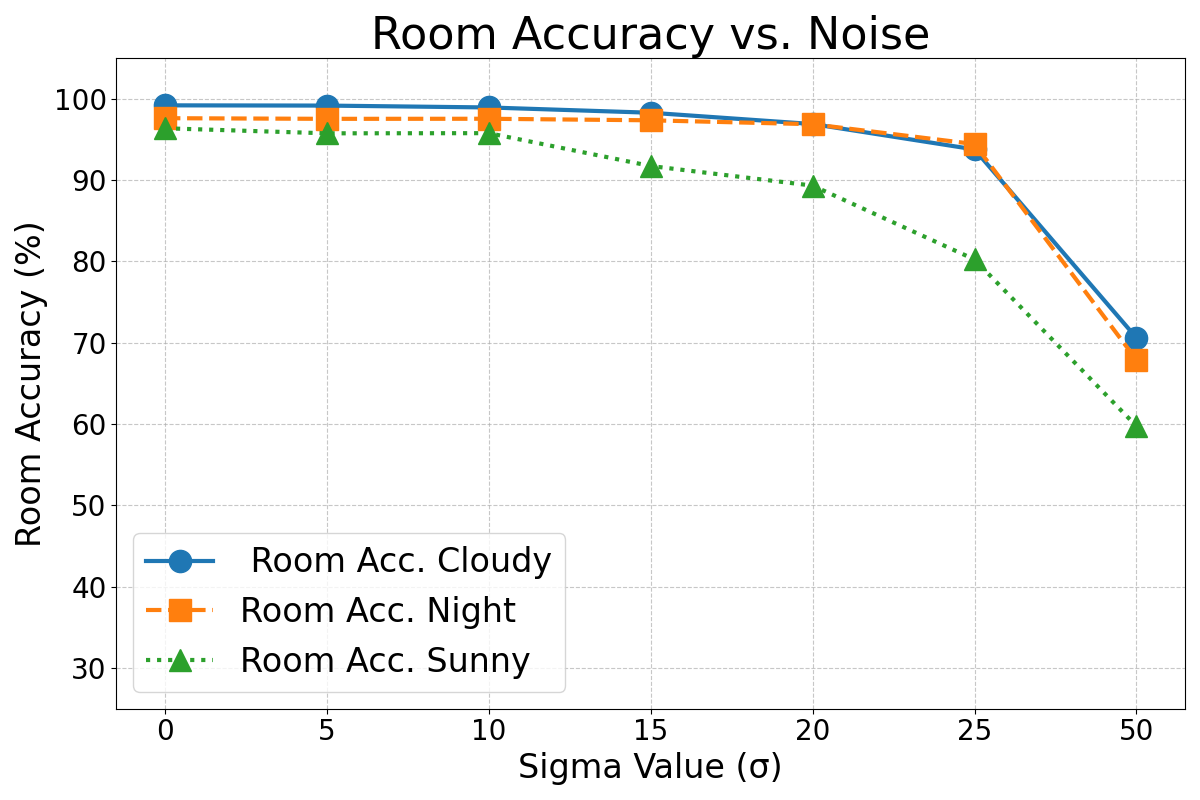}}
      \subfigure[]{\includegraphics[width=0.49\linewidth]{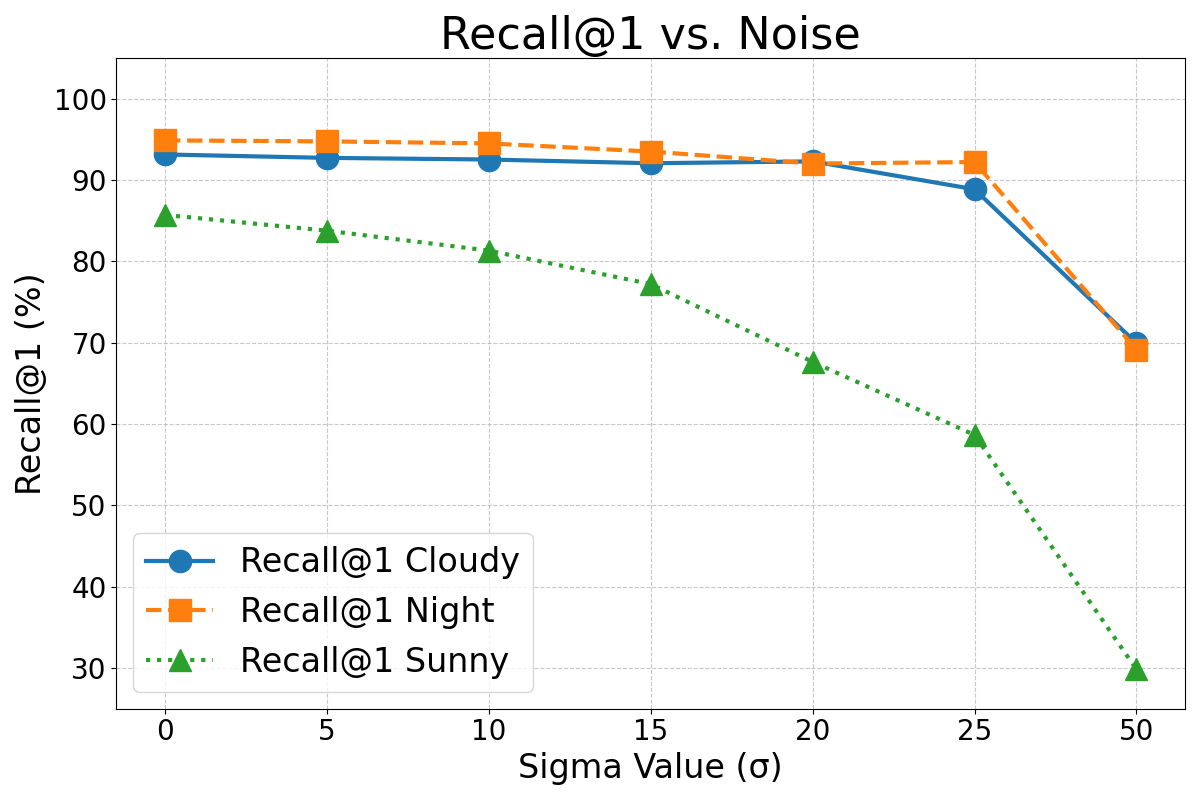}}

 \caption{Performance of the trained models in (a) coarse localization and (b) fine localization with Gaussian noise of different standard deviations $\sigma$.}
    \label{fig:exp2noise}
\end{figure}

Image noise is a frequent artifact in robotic vision. To simulate this, Gaussian noise of standard deviation \(\sigma\) is added to the test images and the images composing the visual map. Figure \ref{fig:exp2noise} shows the average performance degradation for different levels of noise.

\vspace{0.5cm}

\subsubsection{Occlusions effect}

\begin{figure}[!htb]%
    \centering
      \subfigure[]{\includegraphics[width=0.49\linewidth]{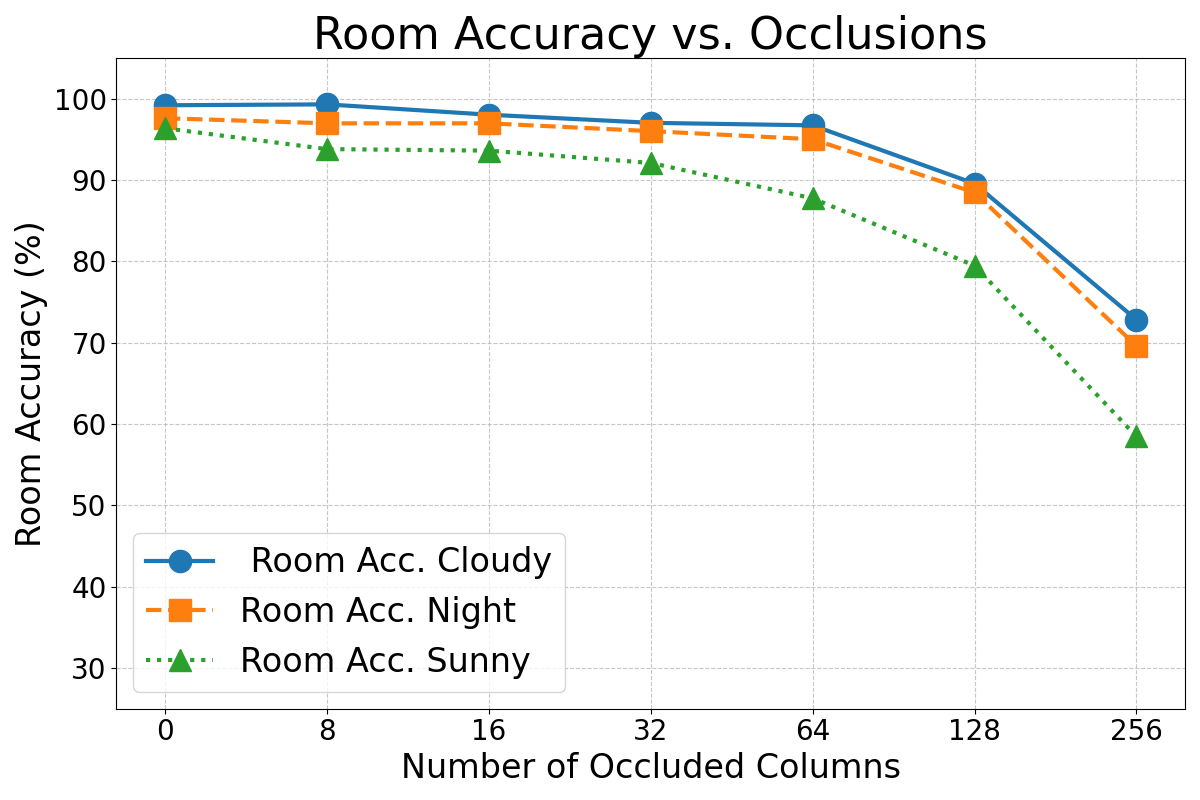}}
      \subfigure[]{\includegraphics[width=0.49\linewidth]{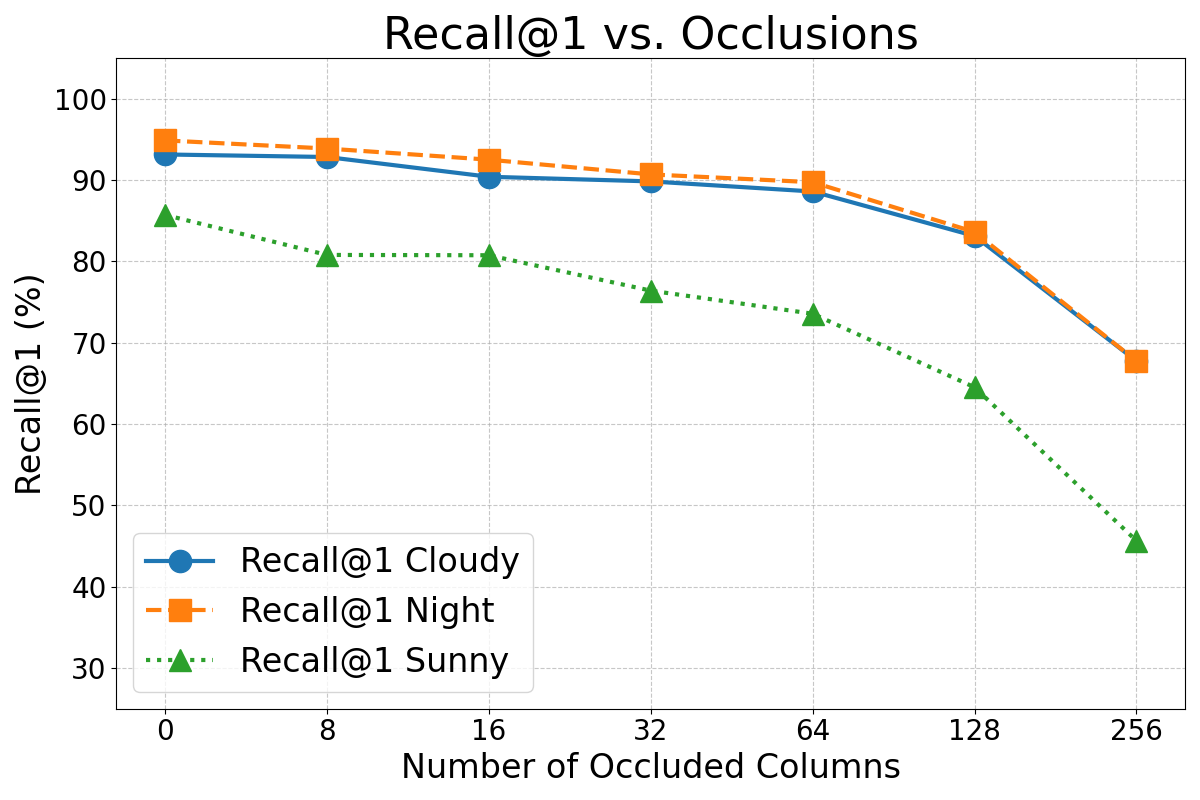}}

 \caption{Performance of the trained models in (a) coarse localization and (b) fine localization with occlusions of different magnitudes.}
    \label{fig:exp2occlusions}
\end{figure}

Mobile robots often encounter scenes where parts of the view are occluded by other objects or people, or even by parts of the robot itself. Occlusions have been simulated by setting a varying number of panoramic image columns to black. Figure \ref{fig:exp2occlusions} illustrates the impact of increasing occlusion on localization performance.

\subsubsection{Motion blur effect}

\begin{figure}[!htb]%
    \centering
      \subfigure[]{\includegraphics[width=0.49\linewidth]{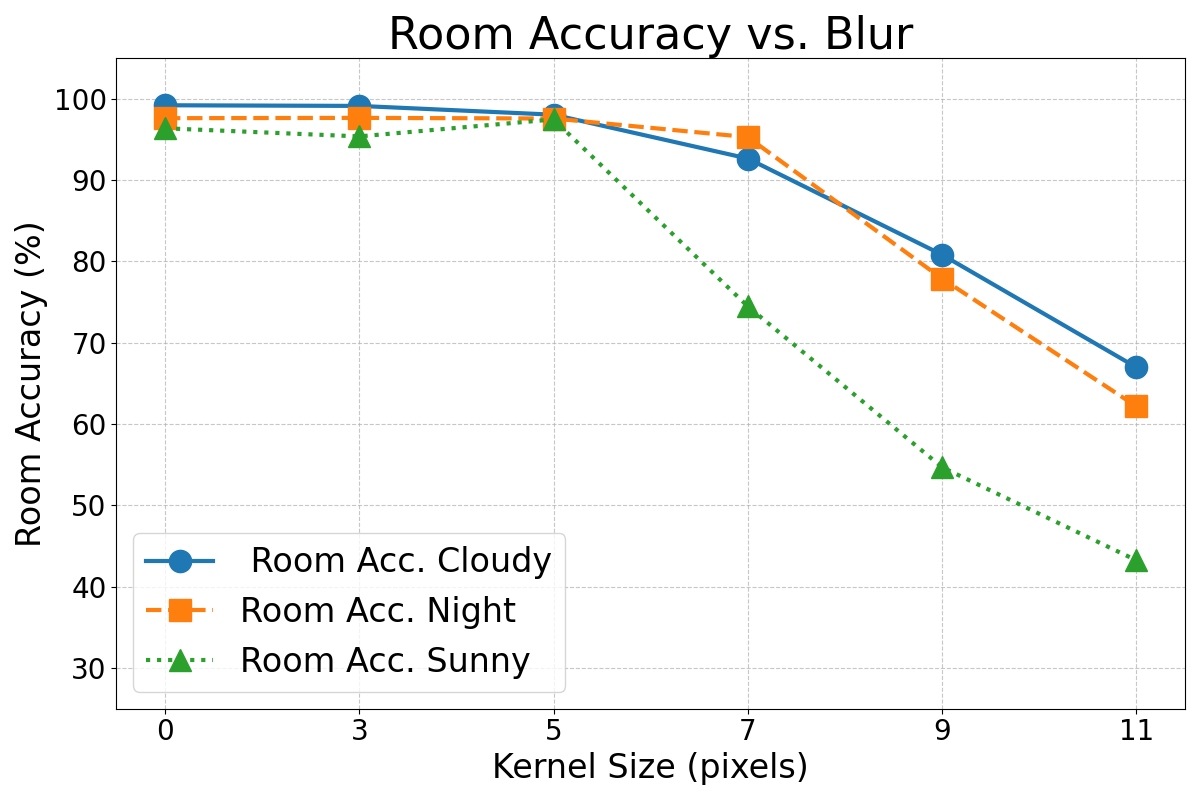}}
      \subfigure[]{\includegraphics[width=0.49\linewidth]{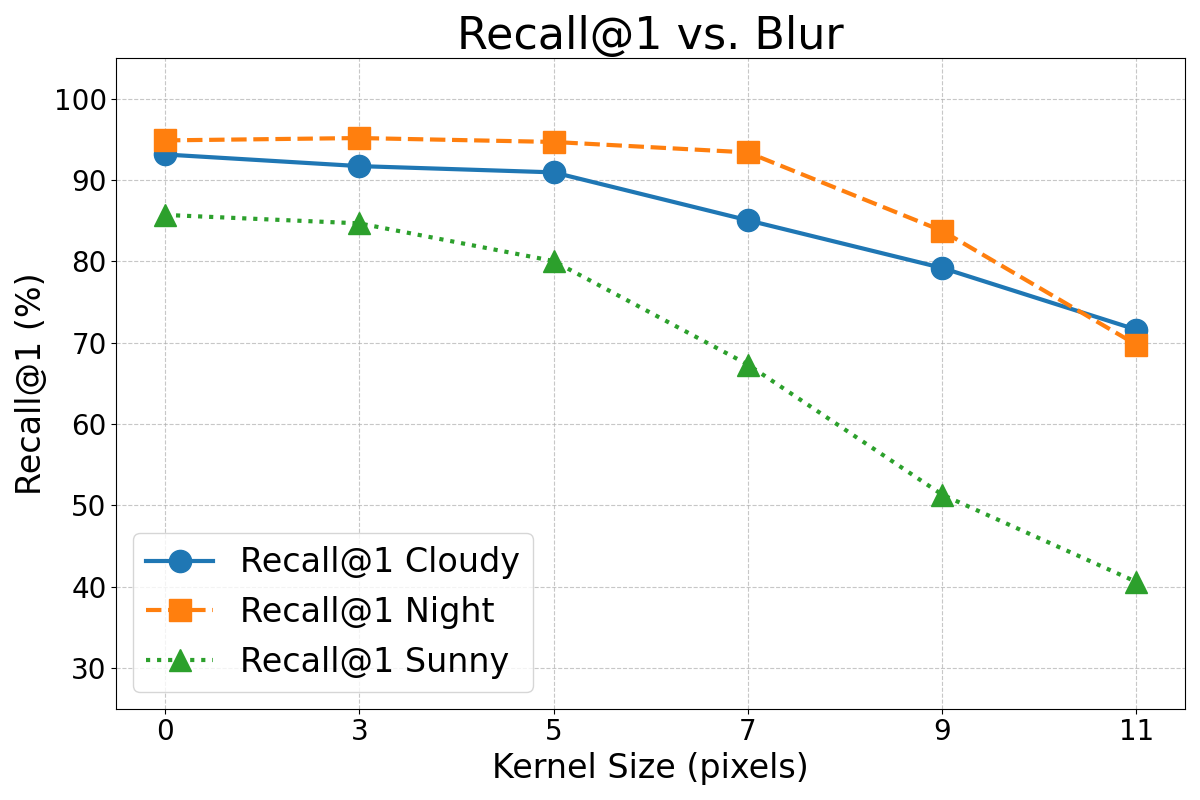}}

 \caption{Performance of the trained models in (a) coarse localization and (b) fine localization with motion blur of different kernel sizes.}
    \label{fig:exp2blur}
\end{figure}

Motion blur occurs when images are captured while the robot is moving or turning. When this happens, the objects of the scene appear blurred. This effect has been simulated by applying a horizontal convolution mask to the images. The severity of the blur was controlled by the size of the mask. Figure \ref{fig:exp2blur} shows the influence of motion blur on the CNN's performance.

\vspace{0.5cm}

\noindent As expected, Figures \ref{fig:exp2noise}, \ref{fig:exp2occlusions}, and \ref{fig:exp2blur} show that the performance of this method degraded as the intensity of these effects increased. However, in general terms, the error rates did not increase substantially until the effects became very pronounced. It is important to note that the magnitude of the simulated effects in this experiment was much larger than the typical variations that may happen in real operation condition and the variations present in the original dataset (see Figure \ref{fig:exp2effects}). For instance, the occlusion caused by the mirror's support structure is a constant and noticeable feature in the images, yet the results from Experiment 1 demonstrate that it does not significantly impair the VPR performance.

\vspace{0.5cm}

\subsection{Experiment 3. Performance study in different environments}

The objective of this experiment is to evaluate the generalization capability of the triplet network-based approach in larger and more varied scenarios with significant environmental changes.  For this purpose, the procedure from Experiment 1 is replicated for three additional environments from the COLD indoor database (FR-B, SA-A and SA-B) and four environments from the 360Loc dataset (atrium, concourse, hall and piatrium). The models are trained using the optimal loss configurations identified in Experiment 1. For the more unstructured outdoor scenarios, which are not composed of rooms, we employed a conventional single-step place recognition approach.

\vspace{0.5cm}

\begin{figure}[!htb]%
    \centering
      \subfigure[]{\includegraphics[width=0.51\linewidth]{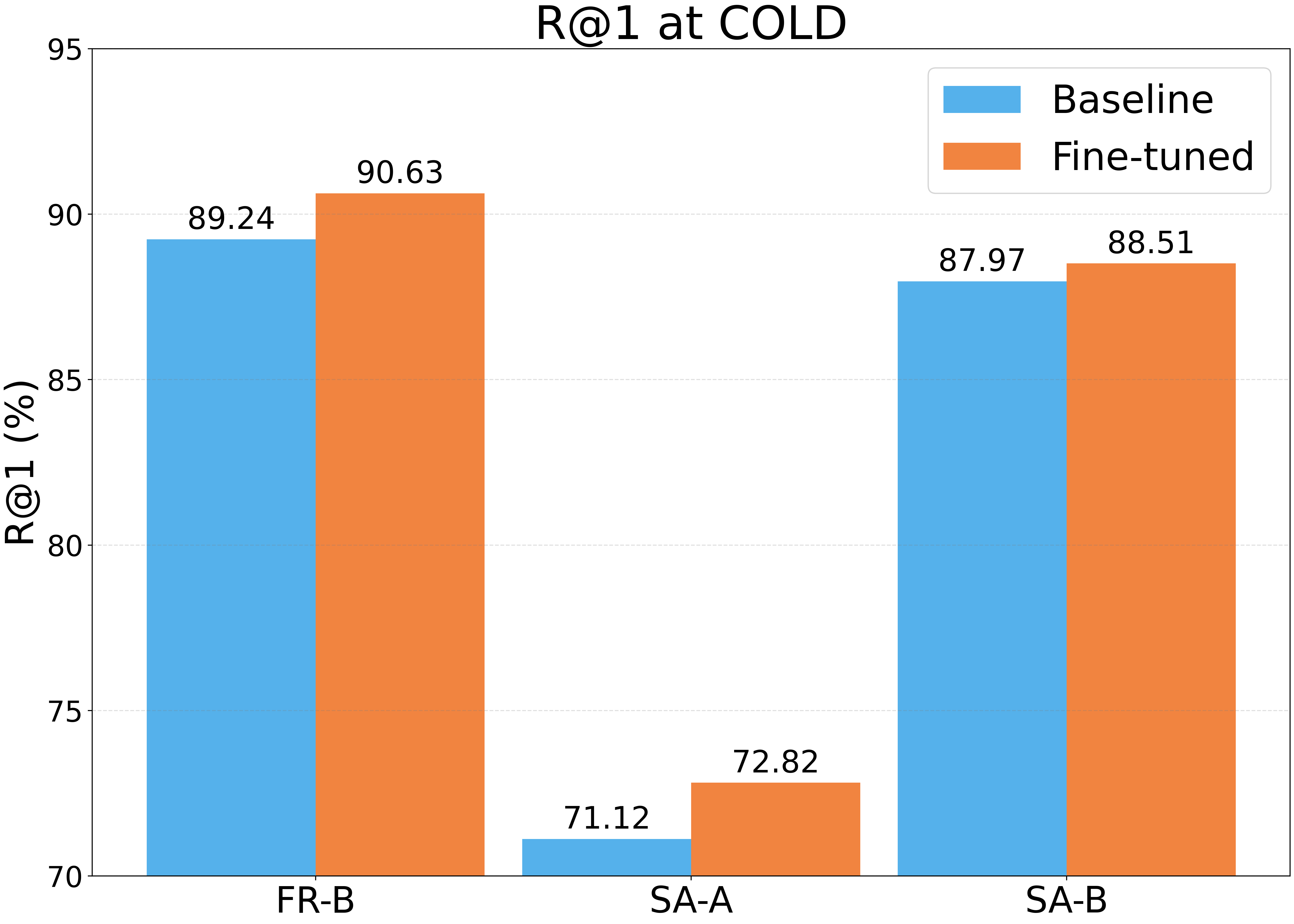}}
      \subfigure[]{\includegraphics[width=0.47\linewidth]{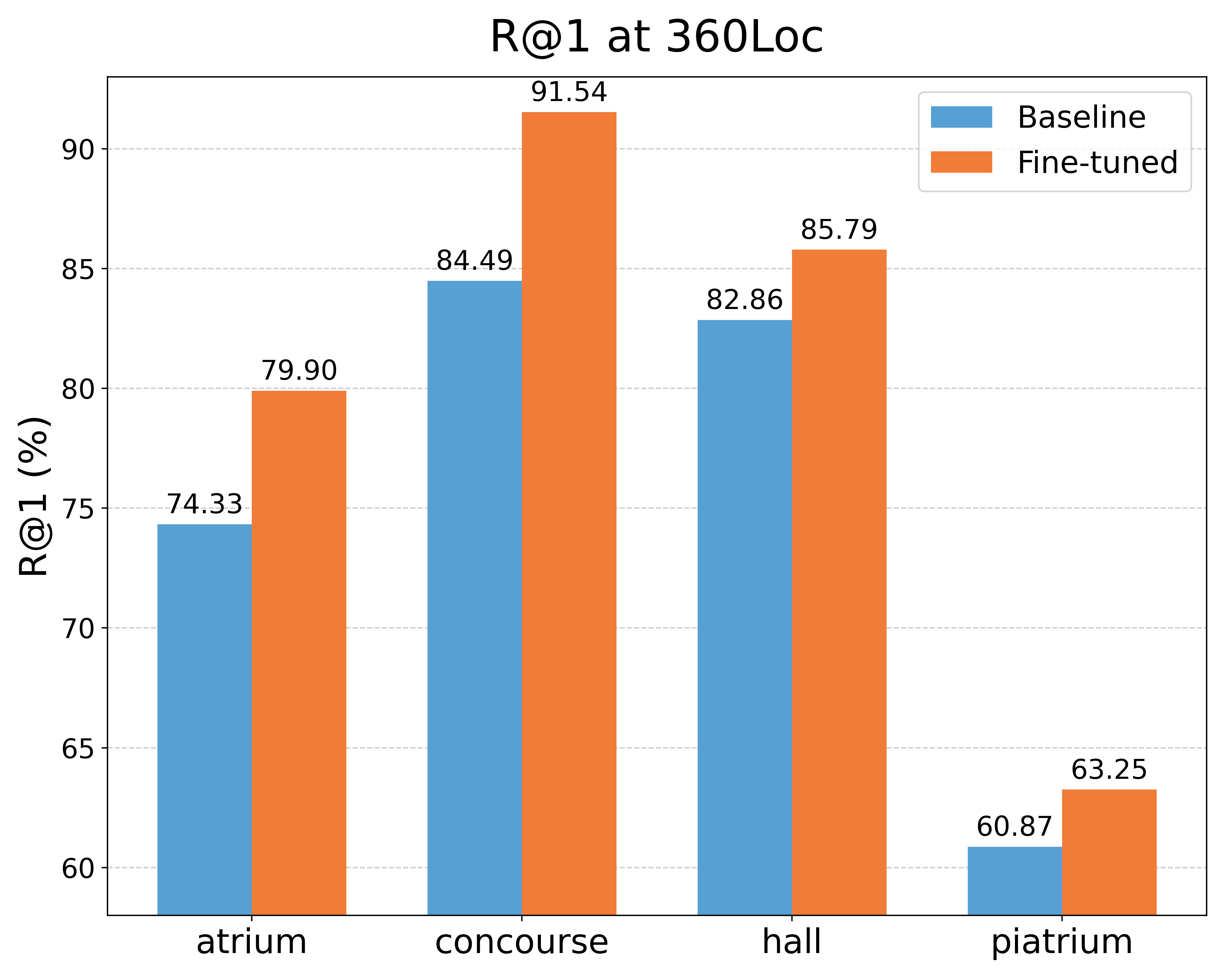}}

 \caption{Performance of the trained models on the (a) COLD and (b) 360Loc databases with respect to the baseline (no fine-tuning).}
    \label{fig:exp3results}
\end{figure}

Tables \ref{tab:exp3imageSetsCOLD} and \ref{tab:exp3imageSets360Loc} detail the number of images in the training and evaluation sets for this experiment. Figure \ref{fig:exp3results} (a) and (b) show the Recall@1 results for the COLD and 360Loc environments, respectively. In this experiment, $d$ was set to 5m for the concourse scenario from the 360Loc database and to 10m for the rest of scenarios from the 360Loc database.

\begin{table}[!htb]
  \centering
  % \resizebox{\linewidth}{!}{
  \begin{tabular}{lcccc}
    \toprule
    Environment & \begin{tabular}[c]{@{}c@{}}Train/ \\ Database (Cloudy) \end{tabular} & \begin{tabular}[c]{@{}c@{}}Test \\ Cloudy\end{tabular} & \begin{tabular}[c]{@{}c@{}}Test \\ Night\end{tabular} & \begin{tabular}[c]{@{}c@{}}Test \\ Sunny\end{tabular} \\
    \midrule
    FR-B & 560 & 2008 & - & 1797 \\
    SA-A & 586 & 2774 & 2267 & - \\
    SA-B & 321 & 836 & 870 & 872 \\
    \bottomrule
  \end{tabular}
  % }
  \caption{Image sets from the COLD database \cite{pronobis2009} used for training and evaluation in Experiment 3.}
  \label{tab:exp3imageSetsCOLD}
\end{table}

\begin{table}[!htb]
  \centering
  % \resizebox{\linewidth}{!}{
  \begin{tabular}{lccc}
    \toprule
    Environment & \begin{tabular}[c]{@{}c@{}}Train/ \\ Database (Day) \end{tabular} & \begin{tabular}[c]{@{}c@{}}Test \\ Day\end{tabular} & \begin{tabular}[c]{@{}c@{}}Test \\ Night\end{tabular} \\
    \midrule
    atrium & 581 & 875 & 1219  \\
    concourse & 491 & 593 & 514 \\
    hall & 540 & 1123 & 1061 \\
    piatrium & 632 & 1008 & 697 \\
    \bottomrule
  \end{tabular}
  % }
  \caption{Image sets from the 360Loc database \cite{huang2024} used for training and evaluation in Experiment 3.}
  \label{tab:exp3imageSets360Loc}
\end{table}

\vspace{0.5cm}

As observed in Figure \ref{fig:exp3results}, the proposed method achieves a notable improvement in performance across all tested scenarios and lighting conditions. Notably, a substantial improvement is observed in the outdoor environments, considering the low number of images used during training. These results underscore the importance of fine-tuning models with omnidirectional images for robust localization in diverse and challenging conditions.

\section{Conclusions}\label{sec:sec5}

This manuscript introduces a hierarchical VPR framework that leverages deep learning models and panoramic images. A central component of our work is an exhaustive evaluation of contrastive triplet loss functions for training these models. The results demonstrate that the proposed loss functions, which incorporate a curriculum learning strategy, significantly outperform existing triplet-based losses at both VPR stages (coarse and fine).

Furthermore, the proposed method was rigorously evaluated under a variety of challenging conditions. It demonstrated robust performance even with a limited number of training images and in the presence of significant visual degradations, including noise, occlusions, and motion blur. Furthermore, its effectiveness was validated across diverse indoor and outdoor environments. In all test cases, this method performed competitively, establishing itself as an accurate, efficient and robust solution for VPR.

Future research will focus on extending the hierarchical approach to large-scale and complex outdoor environments. Additionally, we plan to investigate the integration of other sensory modalities to further enhance place recognition performance and robustness.

\backmatter

% \bmhead{Acknowledgements}

\section*{Declarations}

\begin{itemize}
\item Funding

The Ministry of Science, Innovation and Universities (Spain) has funded this work through FPU23/00587 (M. Alfaro) and FPU21/04969 (J.J. Cabrera). This research work is part of the project PID2023-149575OB-I00 funded by MICIU/AEI/10.13039/501100011033 and by FEDER, UE. It is also part of the project CIPROM/2024/8, funded by Generalitat Valenciana, Conselleria de Educación, Cultura, Universidades y Empleo (program PROMETEO 2025).

\item Conflict of interest/Competing interests 

The authors declare that they have no known competing financial interests or personal relationships that could have appeared to influence the work reported in this paper.

\item Ethics approval and consent to participate

Not applicable

\item Consent for publication

Not applicable

\item Data availability 

 The COLD and 360Loc datasets are public and can be downloaded from their official websites \url{https://www.cas.kth.se/COLD/}, \url{https://github.com/HuajianUP/360Loc}. 

\item Materials availability

Not applicable

\item Code availability 

The code used in the experiments is available at \url{https://github.com/MarcosAlfaro/TripletNetworksIndoorLocalization.git}.

\item Author contribution

Conceptualization: O.R., L.P.; Methodology: J.J., M.F.; Software: M.A.; Validation: J.J., M.F.; Formal analysis: J.J., M.F.; Investigation: M.A., J.J.; Resources: O.R.; Data curation: M.F.; Writing – original draft: M.A., J.J.; Writing – review \& editing: O.R., L.P.; Visualization: M.A., M.F., Supervision: L.P.; Project administration: O.R., L.P.; Funding acquisition: O.R.; L.P.

\end{itemize}

\noindent
% If any of the sections are not relevant to your manuscript, please include the heading and write `Not applicable' for that section. 

%%===================================================%%
%% For presentation purpose, we have included        %%
%% \bigskip command. Please ignore this.             %%
%%===================================================%%
%\bigskip
%\begin{flushleft}%
%Editorial Policies for:

%\bigskip\noindent
%Springer journals and proceedings: \url{https://www.springer.com/gp/editorial-policies}

%\bigskip\noindent
%Nature Portfolio journals: \url{https://www.nature.com/nature-research/editorial-policies}

%\bigskip\noindent
%\textit{Scientific Reports}: \url{https://www.nature.com/srep/journal-policies/editorial-policies}

%\bigskip\noindent
%BMC journals: \url{https://www.biomedcentral.com/getpublished/editorial-policies}
%\end{flushleft}

\end{document}